\documentclass[letterpaper, 10 pt, conference]{ieeeconf}  
\IEEEoverridecommandlockouts                             
\usepackage{graphics} % for pdf, bitmapped graphics files
\usepackage{epsfig} % for postscript graphics files
\usepackage{times} % assumes new font selection scheme installed
\usepackage{amsmath} % assumes amsmath package installed
\usepackage{amssymb}  % assumes amsmath package installed
\usepackage{xcolor}
\usepackage{bm}
\usepackage{cancel}
\usepackage{color}
\usepackage{setspace}
\usepackage{wrapfig}
\usepackage{tikz}
\usetikzlibrary{shapes,arrows.meta,calc}
\usetikzlibrary{positioning}
\usepackage{caption}
\usepackage{subcaption}
\usepackage{booktabs}
\usepackage{multirow}
\usepackage{multicol}
\usepackage{soul}
\usepackage{cite}
\usepackage[normalem]{ulem}
\usepackage{xcolor}

\let\labelindent\relax % Use for IEEE templates to suppress the already defined \labelindent
\usepackage{enumitem}
\usepackage{tabularx}
\usepackage{courier}
\usepackage{algorithm}
\usepackage{algorithmic}

\usepackage{dblfloatfix} % to place table* at bottom of page

\usepackage{pifont}% http://ctan.org/pkg/pifont

\newcommand{\todocite}[1]{\textcolor{red}{[TODO(cite)]}}

\allowdisplaybreaks

\newcommand{\revision}[1]{{\color{black}#1}}
\newcommand{\revisionnew}[1]{{\color{black}#1}}

\newcommand{\norm}[1]{\lVert #1 \rVert}

\newtheorem{remark}{Remark}

\title{\LARGE 
\textbf{
DiffTune: Auto-Tuning through Auto-Differentiation
}
}

\author{Sheng Cheng, Minkyung Kim$^*$, Lin Song$^*$, Chengyu Yang, Yiquan Jin, Shenlong Wang, and Naira Hovakimyan
\thanks{*This work is supported by NASA under the cooperative agreement 80NSSC20M0229,  NSF-AoF Robust Intelligence (2133656), NSF SLES (2331878), Air Force Office of Scientific Research (AFOSR) grant FA9550-21-1-0411, Amazon Research Award, and Illinois-Insper Collaborative Research Fund. 
}
\thanks{\noindent $^*$These authors contributed equally to this work. \newline
S. Cheng, M. Kim, L. Song, C. Yang, Y. Jin, and N. Hovakimyan are with the Department of Mechanical Science and Engineering, and S.~Wang is with the Department of Computer Science. All authors are with the University of Illinois Urbana-Champaign, USA. ({\tt\small email: \{chengs,mk58,linsong2,cy45,yiquanj2,shenlong, nhovakim\}@illinois.edu})}
}

\begin{document}

% make the title area
\renewcommand\thefigure{\arabic{figure}}    

\makeatletter
\g@addto@macro\@maketitle{
\setcounter{figure}{0}
\centering
        \begin{tikzpicture}
    \node (img) {\includegraphics[width=\linewidth]{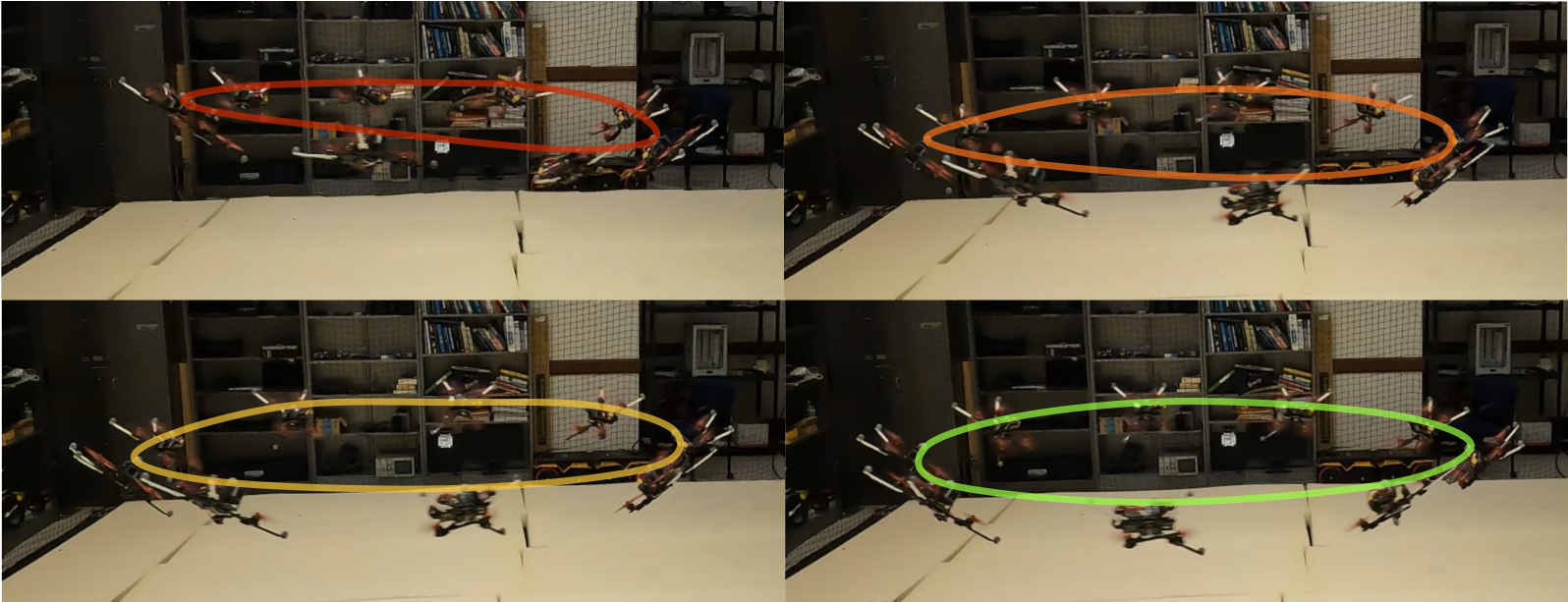}};
    {\color{white}
    \coordinate (trial0) at (-8,3);
    \coordinate (trial3) at (1,3);
    \coordinate (trial6) at (-8,-0.4);
    \coordinate (trial10) at (1,-0.4);
    \node[] at (trial0) {Trial 0};
    \node[] at (trial3) {Trial 3};
    \node[] at (trial6) {Trial 6};
    \node[] at (trial10) {Trial 10};
    }
    {\color{white}
    \coordinate (rmse0) at (-1.5,0.4);
    \coordinate (rmse3) at (7.5,0.4);
    \coordinate (rmse6) at (-1.5,-3);
    \coordinate (rmse10) at (7.5,-3);
    \node[] at (rmse0) {RMSE 0.53 [m]};
    \node[] at (rmse3) {RMSE 0.29 [m]};
    \node[] at (rmse6) {RMSE 0.21 [m]};
    \node[] at (rmse10) {RMSE 0.15 [m]};
    }
\end{tikzpicture}
	\captionof{figure}{Evolution of a quadrotor's tracking performance using DiffTune in 10 trials (with 3.5x reduction on the root-mean-squared-error (RMSE)). The quadrotor is commanded to track a 3 m/s circular trajectory (9 m/s$^2$ centripetal acceleration). The tuning is conducted over 12 parameters. Video recordings of the four trials are available in the supplementary material.}
 \label{fig: exp stacked images 3mps}
	\vspace{-2ex}
}
% \makeatother

\maketitle

% for editing
\thispagestyle{plain}
\pagestyle{plain}
% %% for submission
% \thispagestyle{empty}
% \pagestyle{empty}

% TODOs/notes for the revision:

\begin{abstract}
The performance of robots in high-level tasks depends on the quality of their lower-level controller, which requires fine-tuning. However, the intrinsically nonlinear dynamics and controllers make tuning a challenging task when it is done by hand. 
In this paper, we present DiffTune, a novel, gradient-based automatic tuning framework. We formulate the controller tuning as a parameter optimization problem.
Our method unrolls the dynamical system and controller as a computational graph 
and updates the controller parameters through gradient-based optimization. 
The gradient is obtained using sensitivity propagation, which is the only method for gradient computation when tuning for a physical system instead of its simulated counterpart. Furthermore, we use $\mathcal{L}_1$ adaptive control to compensate for the uncertainties (that unavoidably exist in a physical system) such that the gradient is not biased by the unmodelled uncertainties. 
We validate the DiffTune on a Dubin's car and a quadrotor in challenging simulation environments. In comparison with state-of-the-art auto-tuning methods, DiffTune achieves the best performance in a more efficient manner owing to its effective usage of the first-order information of the system.
Experiments on tuning a nonlinear controller for quadrotor show promising results, where DiffTune achieves 3.5x tracking error reduction on an aggressive trajectory in only 10 trials over a 12-dimensional controller parameter space.
\end{abstract}

\begin{center}
    SUPPLEMENTARY MATERIAL
\end{center}
Video: https://youtu.be/g42UxcIHUdg \\
Code: https://github.com/Sheng-Cheng/DiffTuneOpenSource

\section{Introduction}
Robotic systems are at the forefront of executing intricate tasks, relying on the prowess of their low-level controllers to deliver precise and agile motions. An optimal controller design starts with a meticulous analysis to ensure stability, followed by parameter tuning to achieve the intended performance on real-world robotic platforms. 
Traditionally, controller tuning is done either by hand using trial-and-error or proven methods for specific controllers (e.g., Ziegler–Nichols method for proportional-integral-derivative (PID) controller tuning~\cite{o2009handbook}). Nevertheless, manual tuning often demands seasoned experts and is inefficient, particularly for systems with lengthy loop times or extensive parameter space.

\begin{figure}
    \centering
    \includegraphics[width = \columnwidth]{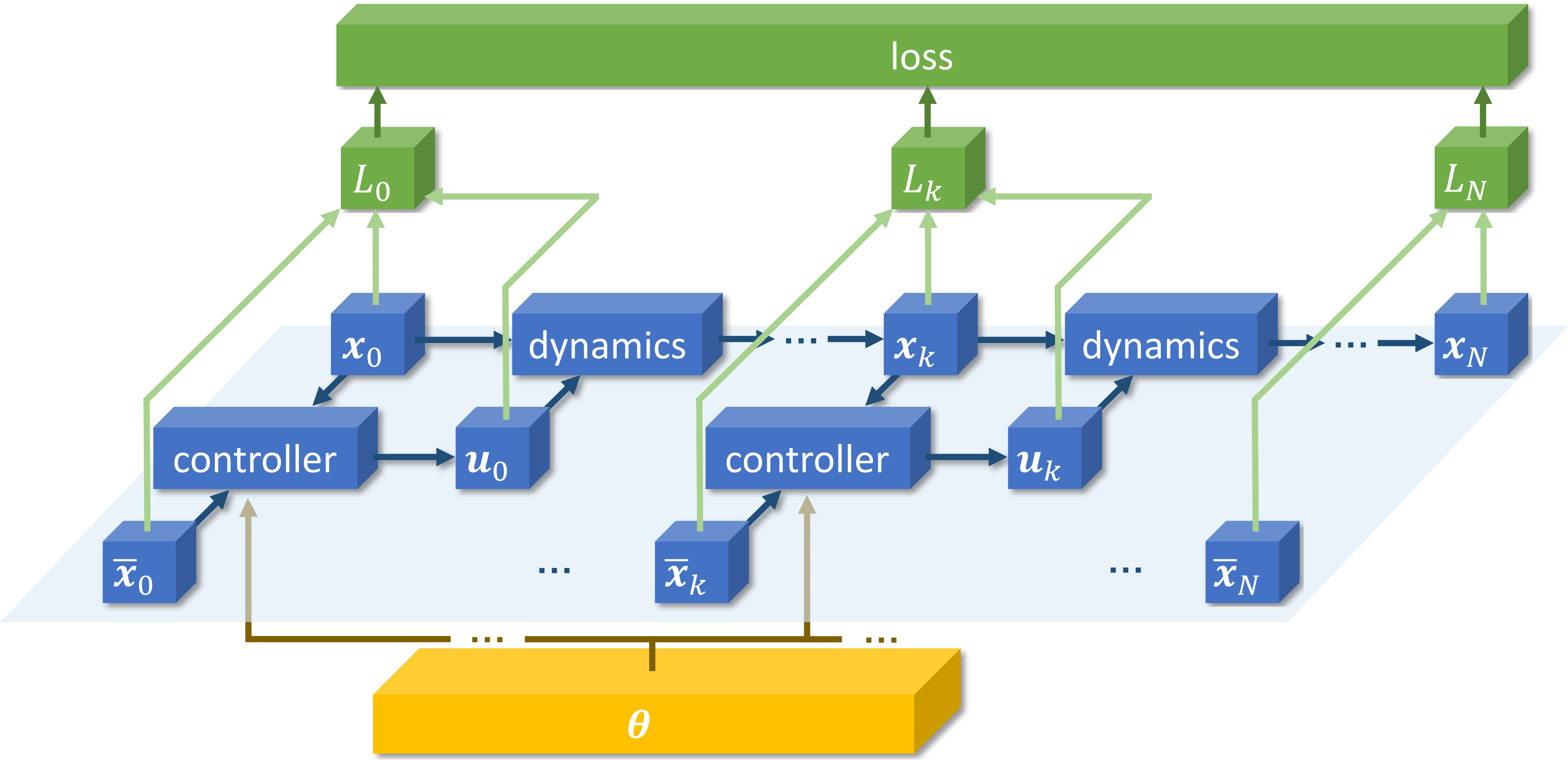}
    \caption{Illustration of an unrolled dynamical system as a computational graph.}
    \label{fig: control system as a computational graph}
    
    \vspace{-0.8cm}
\end{figure}

To improve efficiency and performance, automatic tuning (or auto-tuning) methods have been investigated. Such methods integrate system knowledge, expert experience, and software tools to determine the best set of controller parameters, {especially for the widely used PID controllers~\cite{yu2006autotuning, aastrom1993automatic,li2006patents}}. 
Commercial auto-tuning products have been available since~1980s~\cite{zhuang1993automatic,aastrom1993automatic}.
A desirable auto-tuning scheme should have the following three qualities: i) stability of the target system; ii) compatibility with physical systems' data; and iii) efficiency for online deployment, possibly in real-time. However, how to design the auto-tuning scheme, simultaneously having the above three qualities, for general controllers, is still a challenge.

Existing auto-tuning methods can be categorized into model-based~\cite{trimpe2014self,kumar2021diffloop} and model-free~\cite{marco2016automatic, berkenkamp2016safe,calandra2014bayesian,lizotte2007automatic,loquercio2022autotune, mehndiratta2021can}. 
Both approaches iteratively select the next set of parameters for evaluation that is likely to improve the performance over the previous trials. 
Model-based auto-tuning methods leverage the knowledge of the system model to improve performance, often using the gradient of the performance criterion (e.g., tracking error) and applying gradient descent so that the performance can improve based on the local gradient information~\cite{trimpe2014self,kumar2021diffloop}.
Stability can be ensured by explicitly leveraging knowledge about the system dynamics. However, model-based auto-tuning might not work in a real environment, where the knowledge about the dynamical model might be imperfect. This issue is especially severe when controller parameters are tuned in simulation and then deployed to a physical system.

Model-free auto-tuning methods approximate gradient or a surrogate model to improve the performance. 
Representative approaches include Markov chain Monte Carlo (MCMC)~\cite{loquercio2022autotune}, deep neural network (DNN)~\cite{mehndiratta2021can}, and Bayesian optimization (BO)~\cite{marco2016automatic, berkenkamp2016safe,calandra2014bayesian,lizotte2007automatic,moriconi2020high}. 
Such approaches often make no assumptions about the model and have the advantage of compatibility with physical systems' data owing to their data-driven nature.
\revision{However, it is hard to establish stability guarantees with most of the data-driven methods (e.g., MCMC and DNN), where empirical methods are often applied.
Bayesian optimization has the advantage of establishing stability/safety guarantees, but it can be inefficient when tuning in high-dimensional (e.g., $>$20) parameter spaces. }

To overcome the challenges in the auto-tuning scheme, we present DiffTune: an auto-tuning method based on auto-differentiation (AD). 
Our method is inspired by the ``end-to-end'' idea from the machine learning community. Specifically, in the proposed scheme, the gradient of the loss function (evaluating the performance of the controller) with respect to the controller parameters can be directly obtained and then applied to gradient descent to improve the performance. 
DiffTune is generally applicable to tune all the controller parameters as long as the system dynamics and controller are differentiable (we will define ``differentiable'' in Section~\ref{sec:  formulation}), which is the case with most of the systems. For example, algebraically computed controllers, e.g., with the structure of gain-times-error (PID~\cite{kumar2021diffloop}), are differentiable. Moreover, following the seminal work~\cite{amos2017optnet} that differentiates the \texttt{argmin} operator using the Implicit Function Theorem, one can see that controllers relying on solutions of an optimization problem to generate control actions (e.g., model predictive control (MPC)~\cite{amos2018differentiable,east2020infinite}, optimal control~\cite{jin2020pontryagin,jin2022learning}, safe controllers enabled by control barrier function~\cite{ma2022learning,xiao2021barriernet,xiao2022differentiable,parwana2021recursive}, linear-quadratic regulator (LQR)~\cite{vien2021differentiable}) are also differentiable.

We build DiffTune by unrolling the dynamical system into a computational graph and then applying AD to compute the gradient. Since the structure of the dynamics and controller are untouched by the unrolling operation, the system is still interpretable, which is a distinctive feature compared to the NN-structured dynamics or controllers (widely applied in reinforcement learning). Furthermore, existing tools that support AD (e.g., PyTorch~\cite{NEURIPS2019_9015}, TensorFlow~\cite{tensorflow2015-whitepaper}, JAX~\cite{jax2018github}, and CasADi~\cite{Andersson2019}) can be conveniently applied for gradient computation.

However, when tuning physical systems, we need gradient information based on the data collected from such systems. In this scenario, the computational graph is broken because the states of the system are obtained from sensors rather than by evaluating the dynamics function. The broken graph forbids the usage of AD over the computational graph.
We present an alternative way of gradient computation, called sensitivity propagation, 
which is based on the sensitivity equation~\cite{khalil2015nonlinear} of a dynamical system.
It propagates the sensitivity of the system state to the controller parameters in the forward direction in parallel to the dynamics' propagation. 
% The unique aspect of the sensitivity propagation lies in its capability to incorporate real systems' data, in which case the computational graph is broken because the new system states are obtained through sensor measurements or state estimation instead of evaluating those from the dynamics. 
% On the contrary, the broken computational graph forbids the usage of reverse-mode auto-differentiation, making sensitivity propagation the preferred approach.
Lastly, the gradient of the loss to controller parameters is simply a weighted sum of the sensitivities. Furthermore, uncertainties and disturbances exist in physical systems. If they are not dealt with, the resulting gradient based on the nominal dynamics (free of uncertainty and disturbance) will be biased. We propose to use $\mathcal{L}_1$ adaptive control~\cite{hovakimyan2010L1} to compensate for the uncertainties and disturbances so that the physical system behaves similarly to the nominal system. The uncertainty compensation will preserve the gradient from being biased, thus resulting in more efficient tuning.

DiffTune enjoys the three earlier mentioned qualities simultaneously: \textbf{stability} is inherited from the controllers with stability guarantees by design; \textbf{compatibility with physical systems' data} is enabled by the sensitivity propagation; \textbf{efficiency} is provided since the sensitivity propagation runs forward in time and in parallel to the system's evolution. We have validated DiffTune in both simulations and experiments. DiffTune achieves smaller loss more efficiently than strong baseline auto-tuning methods AutoTune~\cite{loquercio2022autotune} and SafeOpt~\cite{berkenkamp2016safe} (and its variant~\cite{duivenvoorden2017constrained}) in simulations. Notably, in experiments, DiffTune achieves a 3.5x reduction in tracking error in only 10 trials when tuning a nonlinear controller (12-dimensional parameter space) of a quadrotor for tracking an aggressive trajectory, demonstrating the efficacy of the proposed approach.

\textbf{Our contributions are summarized as follows:} i) We propose an auto-tuning method for controller parameters over nonlinear dynamical systems and controllers in general forms by formulating the tuning problem as a parameter optimization problem. Only differentiability of the dynamics, controller, and loss function (for tuning) is required. % We use projected gradient descent to ensure the stability of the system during tuning. 
ii) We treat the unrolled system as a computational graph, over which we use auto-differentiation to compute the gradient efficiently. Specifically, we propose sensitivity propagation, which is compatible with data collected from a physical system and can be efficiently computed online. iii) We combine tuning with the $\mathcal{L}_1$ adaptive control to compensate for the model uncertainties in a physical system that can bias the computed gradient for tuning. iv) We validate the proposed approach in extensive simulations and experiments, where the compatibility with physical data, stability, and efficiency of DiffTune are demonstrated. 

\revision{A previously published paper~\cite{cheng2023difftunePLus} has studied an extension of DiffTune for hyperparameter-free auto-tuning by optimizing the gradient update, with results validated only by simulated systems. }
\revisionnew{In this article, our focus is to introduce DiffTune to solve the auto-tuning problem with a physical system, which has been validated through extensive experiments in Section~\ref{sec: experiments}.
We show that simulation-based auto-tuning can lead to parameter overfitting to a particular simulated system, which does not apply to or even fail the physical system (details in Section~\ref{subsec: sim-based auto-tuning}). To handle the uncertainties that can bias the gradient used in the auto-tuning of a physical system, we use the $\mathcal{L}_1$ adaptive control for uncertainty compensation with the methodology introduced in Section~\ref{subsec: L1 for real system tuning} and experimental results detailed in Section~\ref{subsec: ablation with L1}.}

The remainder of the paper is organized as follows: Sections~\ref{sec: related work} and~\ref{sec: formulation} review related work and background, respectively, of this paper. Section~\ref{sec: method} describes our auto-tuning method and sensitivity propagation. We also discuss uncertainty handling when using physical systems' data. Section~\ref{sec: simulation} shows the simulation results on a Dubins' car and on a quadrotor.
Section~\ref{sec: experiments} demonstrates the experimental results on a quadrotor.
Finally, Section~\ref{sec: conclusion} concludes the paper. 

\section{Related Work}\label{sec: related work}

Our approach closely relates to classical work on automatic parameter tuning and recent learning-based controllers. In this section, we briefly review previous work in the following directions.

\noindent\textbf{Model-based auto-tuning} leverages model knowledge to infer the parameter choice for performance improvement.
In~\cite{trimpe2014self}, an auto-tuning method is proposed for LQR. The gradient of a loss function with respect to the parameterized quadratic matrix coefficients is approximated using Simultaneous Perturbation Stochastic Approximation~\cite{spall2005introduction}, which essentially computes the difference quotients at two random perturbation directions of the current parameter values. 
Simulations and experiments on an inverted pendulum platform demonstrate the effectiveness of this method. 
The authors of~\cite{kumar2021diffloop} applied autodifferentiation to tune a PID controller with input saturation, which is done by differentiating through the model and the feedback loop. Numerical examples are provided to show the efficacy of the proposed method on single-input-single-output systems. 
The authors of~\cite{romero2022model} propose a probabilistic policy search method to efficiently tune a Model Predictive Contouring Control (MPCC) for quadrotor agile flight. The MPCC allows trade-offs between progress maximization and path following in real time, albeit the dimension of parameters grows linearly to the number of gates on a race track. The search of the parameters is turned into maximizing a weighted likelihood function. The approach is validated in real-world scenarios, demonstrating superior performance compared to both manually tuned controllers and state-of-the-art auto-tuning baselines in aggressive quadrotor racing.
In~\cite{romero2023actor}, actor-critic reinforcement learning (RL) is used to tune the parameters of a differentiable MPC for agile quadrotor tracking. The proposed approach enables short-term predictions and optimization of actions based on system dynamics while retaining the end-to-end training benefits and exploratory behavior of an RL agent. Zero-shot transfer to a real quadrotor is demonstrated on different high-level tasks. \revision{The authors of \cite{giordano2018trajectory,srour2023controller} propose to use metrics on the sensitivity of a system's state to uncertain parameters to reduce the variance of the system's performance to uncertainties. The approach has been validated in Monte Carlo simulations to show the benefits of such optimization metrics. 
}

\noindent\textbf{Model-free auto-tuning} relies on a zeroth-order approximate gradient or surrogate performance model to decide the new candidate parameters. 
In~\cite{killingsworth2006pid}, the authors use extremum seeking to sinusoidally perturb the PID gains and then estimate the gradient. Gradient-free methods, e.g., Metropolis-Hastings (M-H) algorithm \cite{hastings1970monte}, have also been used for tuning. 
The M-H algorithm can produce a sequence of random samples from a desired distribution that cannot be directly accessed, whereas a score function is used instead to guide the sampling. In~\cite{loquercio2022autotune}, the M-H algorithm is tailored to tuning the tracking MPC controller for high-speed quadrotor racing that demands minimum-time trajectory completion.
In terms of surrogate models, machine learning tools have been frequently used for their advantages in incorporating data, which, in general, make no assumptions about the systems that produce the data. In \cite{edwards2021automatic}, an end-to-end, data-driven hyperparameter tuning is applied to an MPC using a surrogate dynamical model.
Such a method jointly optimizes the hyperparameters of system identification, task specification, and control synthesis. Simulation validation is conducted in the OpenAI Gym~\cite{openaiGym} environment.

Gaussian Process (GP) is often used as a non-parametric model that approximates an unknown function from input-output pairs with probabilistic confidence measures. This property makes GP a suitable surrogate model that approximates the performance function with respect to the tuned parameters. In~\cite{marco2016automatic}, GP is applied to approximate the unknown cost function using noisy evaluations and then induce the probability distribution of the parameters that minimize the loss. 
The distribution is used to determine new parameters that can maximize the relative entropy, yielding information-efficient exploration of the parameter space. 
% Along with the ongoing exploration 
% The best guess of the optimal parameters can be obtained by checking the maximizer of the induced probability distribution. 
The proposed method is demonstrated to tune an LQR for balancing an inverted pole using a robot arm. 
In~\cite{berkenkamp2016safe}, the authors apply BO~\cite{gelbart2014bayesian} to controller tuning (SafeOpt), which
uses GP to approximate the cost map over controller parameters while constructing safe sets of parameters to ensure safe exploration. 
Safe sets of parameters are constructed while exploring new parameters such that the next evaluation point that can improve the current performance will also be safe. 
Quadrotor experiments are presented where the proposed method is used to tune a PD position controller on a single axis.
Follow-up experiments~\cite{berkenkamp2021bayesian} demonstrate the proposed auto-tuning scheme to quadrotor control with nonlinear tracking controllers. However, one drawback of SafeOpt is that the search for the maximizer requires discretizing the parameters space, which scales poorly to the dimension of parameters. An improved version~\cite{duivenvoorden2017constrained} applies particle swarm heuristics to perform adaptive discretization, which drastically reduces the computation time of SafeOpt to determine the maximizer. \revision{A more recent work~\cite{GIBO} proposes a local BO-like optimizer called Gradient Information with BO (GIBO), which uses a GP for jointly inferring the objective function and its gradients with a probabilistic posterior. The queries of GIBO are chosen to minimize the uncertainty about the gradient. However, the assumption that the objective function is a sample from a known GP prior may not fit the auto-tuning scenario, especially when the system dynamics and the controller hold strong nonlinearities.
}
Bayesian optimization has also been applied to gait optimization for bipedal walking, where GP is used to approximate the cost map of parameterized gaits \cite{calandra2014bayesian,lizotte2007automatic}. 
% Besides GP, DNNs~\cite{mehndiratta2021can} have also been used for model-free tuning. 
% In~\cite{mehndiratta2021can}, a DNN is trained from online-collected data of a quadrotor and then used as a high-fidelity model in simulations to facilitate tuning of a nonlinear MPC. 

\noindent\textbf{Learning for control} % short paragraph
is a recently trending research direction that strives to combine the advantages of model-driven control and data-driven learning for the safe operation of a robotic system. Exemplary approaches include, but not limited to, the following: reinforcement learning \cite{polydoros2017survey,tambon2022certify}, whose goal is to find an optimal policy while gathering data and knowledge of the system dynamics from interactions with the system; imitation learning \cite{ravichandar2020recent}, which aims to mimic the actions taken by a superior controller while making decisions using less information of the system than the superior controller; and iterative learning control \cite{xu2011survey,bristow2006survey}, which constructs the present control action by exploiting every possibility to incorporate past control and system information, typically for systems working in a repetitive mode. A recent survey \cite{brunke2022safe} provides a thorough review of the safety aspect of learning for control in robotics.

\noindent\textbf{Autodifferentiation} % short paragraph
is a technique that evaluates the partial derivative of a function specified by a computer program~\cite{neidinger2010introduction,baydin2018automatic}. Utilizing the inherent nature of computer programs, regardless of their complexity, autodifferentiation capitalizes on the execution of elementary arithmetic operations (such as addition, subtraction, multiplication, division, etc.) and elementary functions (such as exp, log, sin, cos, etc.). By iteratively applying the chain rule to these operations, it automatically computes partial derivatives of any desired order with high accuracy while incurring only a marginal increase in the number of arithmetic operations compared to the original program. Autodifferentiation has a significant advantage over other differentiation methods, e.g., manual differentiation (prone to errors and time-consuming), numerical differentiation like finite difference (poor scalability to high-dimensional inputs and proneness to round-off errors), or symbolic differentiation (suffering from overly complicated symbolic representations of the derivative, known as ``expression swell''). A comprehensive comparison of these differentiation methods is provided in~\cite{baydin2018automatic}. There are two modes of AD, reverse mode and forward mode, both relying on the chain rule to propagate the derivative. The reverse-mode AD requires a forward pass of the computational graph and keeps the values of the intermediate nodes in the memory. Subsequently, a backward pass propagates the partial derivatives from the output to the input of the graph. The forward-mode AD propagates the partial derivatives while conducting the forward pass on the graph, storing both the values and partial derivatives of the intermediate nodes in the memory.

\section{Problem Formulation} \label{sec: formulation}

Consider a discrete-time dynamical system
\begin{equation}\label{eq: dynamics}
    \boldsymbol{x}_{k+1} = f(\boldsymbol{x}_k,\boldsymbol{u}_k), 
\end{equation}
where $\boldsymbol{x}_k \in \mathbb{R}^n$ and $\boldsymbol{u}_k \in \mathbb{R}^m$ are the state and control, respectively, and the initial state $\boldsymbol{x}_0$ is known. The control is generated by a feedback controller that tracks a desired state $\bar{\boldsymbol{x}}_{k} \in \mathbb{R}^n$ such that
\begin{equation}
    \boldsymbol{u}_k = h(\boldsymbol{x}_k,\bar{\boldsymbol{x}}_{k},\boldsymbol{\theta}), \label{eq: feedback controller}
\end{equation}
where $\boldsymbol{\theta} \in \mathbb{R}^p$ denotes the parameters of the controller, e.g., $\boldsymbol{\theta} \in \mathbb{R}^2$ may represent the P- and D-gain in a PD controller. We assume that the state $\boldsymbol{x}_k$ can be measured directly or, if not, an appropriate state estimator is used. Furthermore, we assume the dynamics \eqref{eq: dynamics} and controller \eqref{eq: feedback controller} are differentiable, i.e., the Jacobians $\nabla_{\boldsymbol{x}} f$, $\nabla_{\boldsymbol{u}} f$, $\nabla_{\boldsymbol{x}} h$, and $\nabla_{\boldsymbol{\theta}} h$ exist, which widely applies to general systems.

The tuning task adjusts $\boldsymbol{\theta}$ to minimize an evaluation criterion, denoted by $L(\cdot)$, which is a differentiable function of the desired states $\bar{\boldsymbol{x}}$, actual states $\boldsymbol{x}$, and control ations $\boldsymbol{u}$ over a time interval of length $N$. An illustrative example is the tracking error plus control-effort penalty, where $L(\boldsymbol{x}_{0:N},\bar{\boldsymbol{x}}_{0:N},\boldsymbol{u}_{0:N-1};\boldsymbol{\theta}) = \sum_{k=0}^{N} \norm{\boldsymbol{x}_k -\bar{\boldsymbol{x}}_{k}}^2 + \sum_{k=0}^{N-1}\lambda \norm{\boldsymbol{u}_k}^2$ 
with $\lambda>0$ being the penalty coefficient. We will use the short-hand notation $L(\boldsymbol{\theta})$ for conciseness in the rest of the paper.

With the setup introduced above, controller tuning can be formulated as a parameter optimization problem as follows:
\begin{equation}\label{prob: controller tuning as a parameter optimization}
    \begin{aligned}
    % & \underset{\boldsymbol{\theta} \in \Theta}{\text{minimize}} && L(\boldsymbol{x}_{0:N},\bar{\boldsymbol{x}}_{0:N},\boldsymbol{u}_{0:N-1};\boldsymbol{\theta})\\
    & \underset{\boldsymbol{\theta} \in \Theta}{\text{minimize}} && L(\boldsymbol{\theta})\\
    & \text{subject to} && \boldsymbol{x}_{k+1} = f(\boldsymbol{x}_k,\boldsymbol{u}_k),\\
    & && \boldsymbol{u}_k = h(\boldsymbol{x}_k,\bar{\boldsymbol{x}}_{k},\boldsymbol{\theta}),\\
    & && k \in \{0,1,\dots,N-1\}.
    \end{aligned}
    \tag{P}
\end{equation}
Note that problem~\eqref{prob: controller tuning as a parameter optimization} searches for controller parameter $\boldsymbol{\theta}$ to minimize the loss $L$ subject to the system's dynamics and a chosen controller (to be tuned). Problem \eqref{prob: controller tuning as a parameter optimization} is generally nonconvex due to the nonlinearity in dynamics~$f$ and controller~$h$. We will introduce our method, \textit{DiffTune}, in Section~\ref{sec: method} for auto-tuning, especially for tuning a controller for a physical system. 

\section{Method} \label{sec: method}
We use a gradient-based method to solve problem~\eqref{prob: controller tuning as a parameter optimization} due to its nonconvexity, where the system performance is gradually improved by adjusting the controller parameters using gradient descent.
We unroll the dynamical system~\eqref{eq: dynamics} and controller~\eqref{eq: feedback controller} into a computational graph. Figure~\ref{fig: control system as a computational graph} illustrates the unrolled system, which stacks the iterative procedure of state update via the ``dynamics'' and control-action generation via the ``controller.''
The gradient $\nabla_{\boldsymbol{\theta}}L$ is then applied to update the parameters $\boldsymbol{\theta}$. Specifically, since the parameters are usually confined to a feasible set $\Theta$, we use the projected gradient descent~\cite{parikh2014proximal} to update $\boldsymbol{\theta}$:
\begin{equation}\label{eq: projected gradient descent to update the parameters}
    \boldsymbol{\theta} \leftarrow \mathcal{P}_{\Theta} ( \boldsymbol{\theta} - \alpha \nabla_{\boldsymbol{\theta}}L),
\end{equation}
where $\mathcal{P}_{\Theta}$ is the projection operator that projects its operand into the set $\Theta$, and $\alpha$ is the learning rate. The feasible set $\Theta$ is used here to ensure the stability of the system, where $\Theta$ can be determined via the Lyapunov analysis or empirically determined by engineering practice.

What remains to be done is to compute the gradient $\nabla_{\boldsymbol{\theta}} L $, for which AD can be used when the computational graph is complete (e.g., in simulations).
% Our method is inspired by the backward propagation used in training a neural network (NN): once the structure of the NN and loss function are defined, then the parameters of the NN are updated via gradient descent. Denote the NN parameters by $\boldsymbol \phi$ and the loss function by $l(\boldsymbol{\phi})$. Backward propagation, also known as the reverse-mode auto-differentiation on a computational graph, is nowadays used to obtain the gradient $\nabla_{\boldsymbol{\phi}} l$. Likewise, the computational graph in Fig.~\ref{fig: control system as a computational graph} has the controller parameters $\boldsymbol{\theta}$ as the leaf node and the loss $L(\boldsymbol{\theta})$ to be the root node, whereas all the intermediate states and control actions are non-leaf nodes.
% The computational graph has to be propagated forward first, with the value of each non-leaf node stored in the memory. Then the backward propagation uses chain rule and the stored non-leaf nodes in the memory to trace the graph from root to leaves and compute the desired gradient $\nabla_{\boldsymbol{\theta}} L $. 
AD can be conveniently implemented using off-the-shelf tools like PyTorch~\cite{NEURIPS2019_9015}, TensorFlow~\cite{tensorflow2015-whitepaper}, JAX~\cite{jax2018github}, or CasADi~\cite{Andersson2019}: one will program the computational graph using the dynamics and controller and set the parameter $\boldsymbol{\theta}$ with respect to which the loss function will be differentiated.

However, AD methods cannot incorporate data (state and control) from a physical system because AD relies on a complete computation graph, whereas the computational graph corresponding to a physical system is broken. Specifically, the dynamics \eqref{eq: dynamics} have to be evaluated each time to obtain a new state, which is not the case in a physical system: the states are obtained through sensor measurements or state estimation rather than evaluating the dynamics (see the comparison in Figs.~\ref{fig: conceptual graph} and~\ref{fig: graph with a physical system}). This explains why the computational graph is broken when considered for a physical system.
Thus, AD can only be applied to auto-tuning in simulations, forbidding its usage with physical systems' data. We introduce sensitivity propagation next to address the compatibility with physical systems' data.

\subsection{Sensitivity propagation}
We first break down the gradient $\nabla_{\boldsymbol{\theta}} L $ using chain rule: 
\begin{equation}
    \nabla_{\boldsymbol{\theta}}L =\sum_{k=1}^N \frac{\partial L}{\partial \boldsymbol{x}_k} \frac{\partial \boldsymbol{x}_k}{\partial \boldsymbol{\theta}} + \sum_{k=0}^{N-1} \frac{\partial L}{\partial \boldsymbol{u}_k} \frac{\partial \boldsymbol{u}_k}{\partial \boldsymbol{\theta}}.\label{eq: decomposition of the target derivative to accepting sensitivity propagation}
\end{equation}
Since $\partial L / \partial \boldsymbol{x}_k$ and $\partial L / \partial \boldsymbol{u}_k$ can be determined once $L$ is chosen, what remains to be done is to obtain $\partial \boldsymbol{x}_k / \partial \boldsymbol{\theta}$ and $\partial \boldsymbol{u}_k / \partial \boldsymbol{\theta}$. Given that the system states $\boldsymbol{x}_k$ are iteratively defined using the dynamics~\eqref{eq: dynamics}, we can derive an iterative formula for $\partial \boldsymbol{x}_k / \partial \boldsymbol{\theta}$ and $\partial \boldsymbol{u}_k / \partial \boldsymbol{\theta}$ by taking partial derivative with respect to $\boldsymbol{\theta}$ on both sides of the dynamics~\eqref{eq: dynamics} and controller~\eqref{eq: feedback controller}:
\begin{subequations}\label{eq: sensitivity propagation}
\begin{align}
    \frac{\partial \boldsymbol{x}_{k+1}}{\partial \boldsymbol{\theta}} = & (\nabla_{\boldsymbol{x}_k} f + \nabla_{\boldsymbol{u}_k} f \nabla_{\boldsymbol{x}_k} h) \frac{\partial \boldsymbol{x}_k}{\partial \boldsymbol{\theta}} + \nabla_{\boldsymbol{u}_k} f \nabla_{\boldsymbol{\theta}} h, \label{eq: iterative Jacobian of state wrt parameter} \\
    \frac{\partial \boldsymbol{u}_k}{\partial \boldsymbol{\theta}} = & \nabla_{\boldsymbol{x}_k} h \frac{\partial \boldsymbol{x}_k}{\partial \boldsymbol{\theta}} + \nabla_{\boldsymbol{\theta}} h, \label{eq: iterative Jacobian of control wrt parameter}
\end{align}
\end{subequations}
with $\partial \boldsymbol{x}_0/\partial \boldsymbol{\theta} = 0$.
Note that \eqref{eq: iterative Jacobian of state wrt parameter} is essentially the sensitivity equation of a system~\cite[Chapter~3.3]{khalil2015nonlinear}. We name the Jacobians $\partial \boldsymbol{x}_k / \partial \boldsymbol{\theta}$ and $\partial \boldsymbol{u}_k / \partial \boldsymbol{\theta}$ by sensitivity states.

\begin{figure}
\begin{subfigure}{\columnwidth}
    \centering
    \includegraphics[width = \columnwidth]{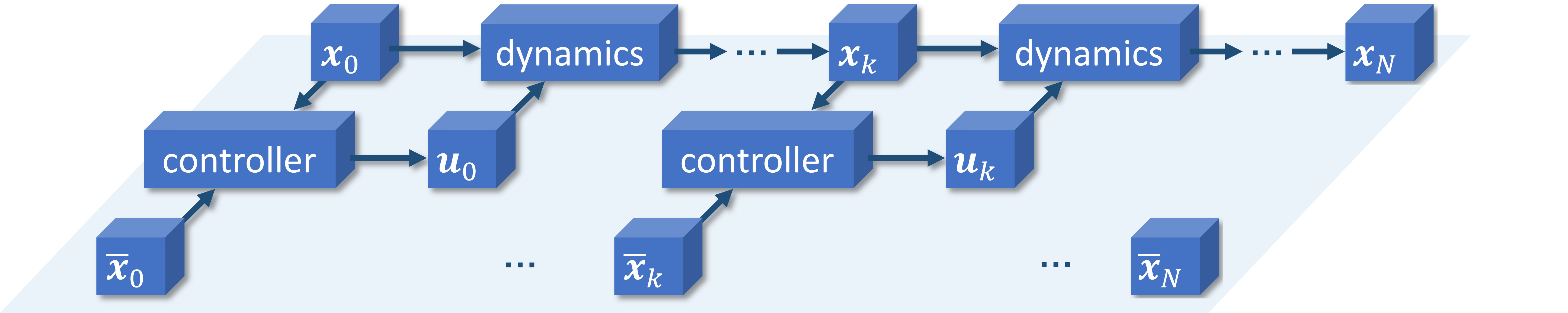}
    \caption{}
    \label{fig: conceptual graph}
\end{subfigure}
\begin{subfigure}{\columnwidth}
    \centering
    \includegraphics[width = \columnwidth]{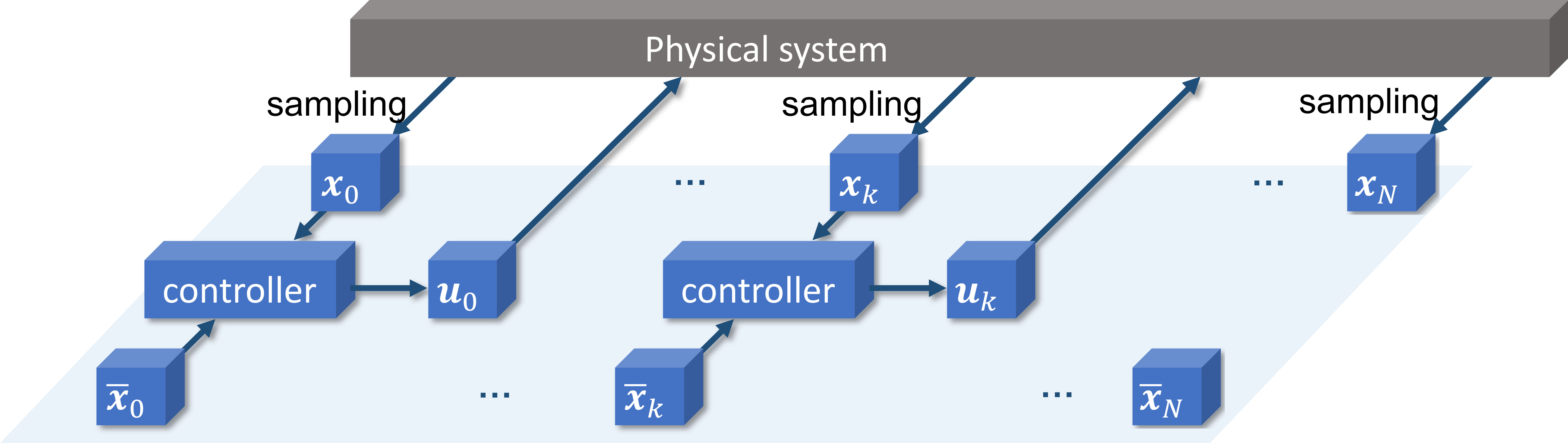}
    \caption{}
    \label{fig: graph with a physical system}
\end{subfigure}
\begin{subfigure}{\columnwidth}
    \centering
    \includegraphics[width = \columnwidth]{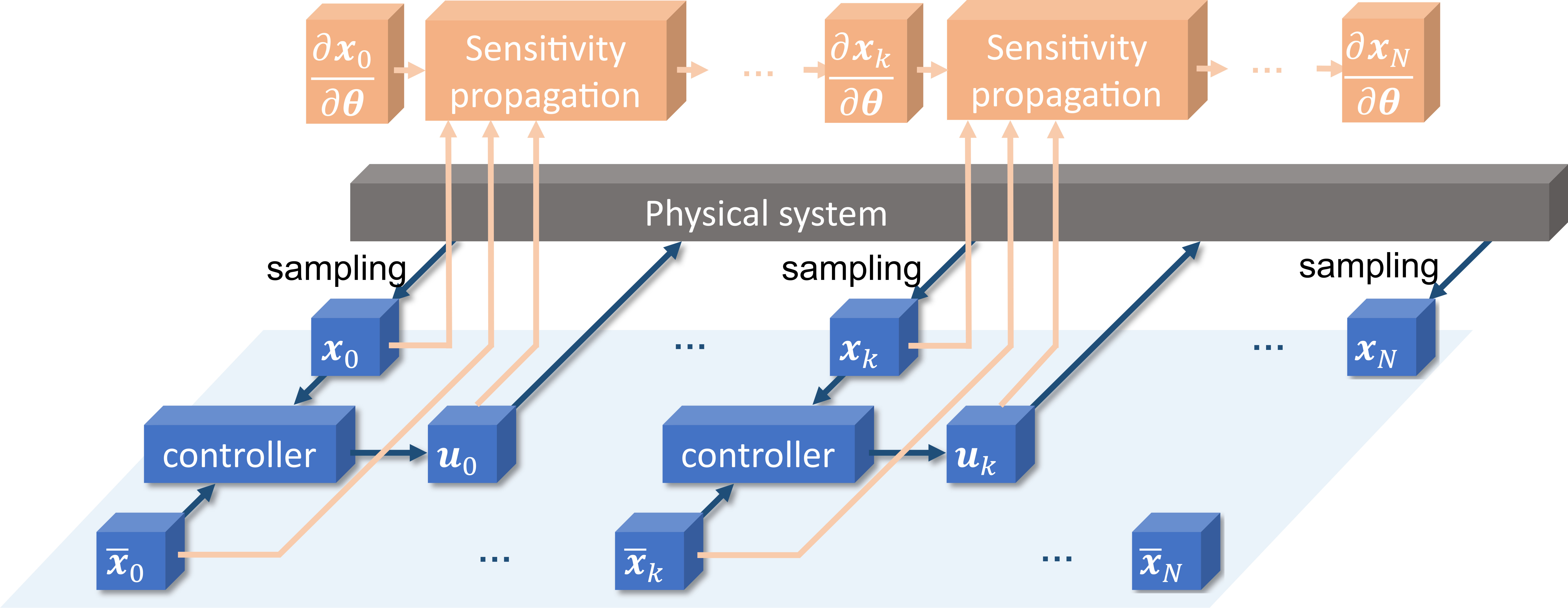}
    \caption{}
    \label{fig: graph with physical system and sensitivity propagation}
\end{subfigure}
\caption{Difference between a conceptual dynamical system~(a) and a physical system~(b). Since a physical system in~(b) does not have a complete computational graph for autodifferentiation, sensitivity propagation is used for gradient computation and is illustrated in (c).}
\label{fig: comparison between different schemes}
\end{figure}

The sensitivity propagation~\eqref{eq: iterative Jacobian of state wrt parameter} works by propagating the sensitivity state $\partial \boldsymbol{x}_k / \partial \boldsymbol{\theta}$ forward in time. In fact, \eqref{eq: iterative Jacobian of state wrt parameter} is a time-varying linear system with the sensitivity state $\partial \boldsymbol{x}_k / \partial \boldsymbol{\theta}$. The system matrix $\nabla_{\boldsymbol{x}_k} f + \nabla_{\boldsymbol{u}_k} f \nabla_{\boldsymbol{x}_k} h$ and the excitation $\nabla_{\boldsymbol{u}_k} f \nabla_{\boldsymbol{\theta}} h$ are computed each time with the data sampled from the physical system. Specifically, the coefficients $\nabla_{\boldsymbol{x}_k} f$, $\nabla_{\boldsymbol{u}_k} f$, $\nabla_{\boldsymbol{x}_k} h$, and $\nabla_{\boldsymbol{\theta}} h$, whose formula are known since $f$ and $h$ are known, are evaluated at sampled state $\boldsymbol{x}_k$ and control $\boldsymbol{u}_k$. Once $\{\partial \boldsymbol{x}_k / \partial \boldsymbol{\theta} \}_{k=0:N}$ and $\{\partial \boldsymbol{u}_k / \partial \boldsymbol{\theta} \}_{k=0:N-1}$ are all computed, $\nabla_{\boldsymbol{\theta}}L$ can be computed as the weighted sum of the sensitivity states, where the weights $\{\partial L / \partial \boldsymbol{x}_k \}_{k=0:N}$ and  $\{\partial L / \partial \boldsymbol{u}_k \}_{k=0:N-1}$ (whose formula are also known) are evaluated at the sampled data. An illustration of how sensitivity propagation works is shown in Fig.~\ref{fig: graph with physical system and sensitivity propagation}.
Furthermore, sensitivity propagation permits online tuning. Since the formulas of $\nabla_{\boldsymbol{x}_k} f$, $\nabla_{\boldsymbol{u}_k} f$, $\nabla_{\boldsymbol{x}_k} h$, and $\nabla_{\boldsymbol{\theta}} h$ can be derived offline, the sensitivity propagation can update $\partial \boldsymbol{x}_{k+1} / \partial \boldsymbol{\theta}$ online whenever the system data $\boldsymbol{x}_k$ and $\boldsymbol{u}_k$ are sampled. Owing to the forward-in-time nature of sensitivity propagation, the horizon $N$ can be adjusted online by need, which further contributes to the flexibility of gradient computation to a varying horizon using sensitivity propagation. We summarize the DiffTune algorithm with sensitivity propagation in Alg.~\ref{algo: full algorithm using sensitivity propatation}.

\begin{remark}
    The sensitivity propagation and forward-mode AD share the same formula as in \eqref{eq: decomposition of the target derivative to accepting sensitivity propagation} and \eqref{eq: sensitivity propagation}. The difference lies in what type of data is applied. Two types of data are considered: The first type is from simulation, where $\boldsymbol{x}_k$ is obtained by the computation $\boldsymbol{x}_k = f(\boldsymbol{x}_{k-1},\boldsymbol{u}_{k-1})$; the second type is sampled from a physical system, where $\boldsymbol{x}_k$ is obtained by either sensor measurements or state estimation, which cannot be represented as the evaluation a mathematical expression. In principle, forward-mode AD can only work with data of the first type, which limits its application to auto-tuning in simulations only. Sensitivity propagation, however, can work with \textit{both} types of data. The most significant usage is associated with the second-type data, which provides a straightforward method of tuning a physical system. When applied to the first type of data, the sensitivity propagation is equivalent to the forward-mode AD.
\end{remark}

\begin{remark}
    One may notice that the sensitivity state $\partial \boldsymbol{x}_0 / \partial \boldsymbol{\theta}$ is set to a zero matrix. This initialization relates to how the sensitivity state is interpreted. Suppose we have sampled the sequence of state $\{\boldsymbol{x}_k\}_{k=0:N}$ and control $\{\boldsymbol{u}_k\}_{k=0:N-1}$ subject to a certain parameter $\boldsymbol{\theta}$. Consider a small perturbation $\boldsymbol{\epsilon} \in \mathbb{R}^p$ to $\boldsymbol{\theta}$. The sensitivity states allow inferring about the state and control sequence subject to the parameter being $\boldsymbol{\theta}+\boldsymbol{\epsilon}$ using first-order approximation (without implementing the controller with the parameter  $\boldsymbol{\theta}+\boldsymbol{\epsilon}$ and then sampling the data). Specifically, we have
    \begin{align}
    \boldsymbol{x}_k(\boldsymbol{\theta}+\boldsymbol{\epsilon}) \approx & \ \boldsymbol{x}_k(\boldsymbol{\theta}) + \frac{\partial \boldsymbol{x}_k}{\partial \boldsymbol{\theta}} \boldsymbol{\epsilon}, \\ 
    \boldsymbol{u}_k(\boldsymbol{\theta}+\boldsymbol{\epsilon}) \approx & \ \boldsymbol{u}_k(\boldsymbol{\theta}) + \frac{\partial \boldsymbol{u}_k}{\partial \boldsymbol{\theta}} \boldsymbol{\epsilon}.
    \end{align}
The sensitivity state $\partial \boldsymbol{x}_k / \partial \boldsymbol{\theta}$ is initialized at zero such that $\boldsymbol{x}_0(\boldsymbol{\theta}+\boldsymbol{\epsilon}) = \boldsymbol{x}_0(\boldsymbol{\theta})$ to ensure the same initial state despite parameter change. Therefore, how the state $\boldsymbol{x}_k(\boldsymbol{\theta} + \boldsymbol{\epsilon})$ will change subject to the $\boldsymbol{\epsilon}$ parameter perturbation can be inferred from the sensitivity $\partial \boldsymbol{x}_k / \partial \boldsymbol{\theta}$ which simply evolves with the sensitivity equation. Furthermore, the sensitivity states allow for auto-tuning without hyperparameters (e.g., learning rate $\alpha$), which is detailed in~\cite{cheng2023difftunePLus}.
    
\end{remark}

\begin{remark}
    Although AD cannot be applied to the entire computational graph when using data from a physical system, it can still be applied to obtain the Jacobians $\nabla_{\boldsymbol{x}_k} f$, $\nabla_{\boldsymbol{u}_k} f$, $\nabla_{\boldsymbol{x}_k} h$, and $\nabla_{\boldsymbol{\theta}} h$. Since the iterative structure in~\eqref{eq: sensitivity propagation} remains the same among iterations, AD packages like PyTorch~\cite{NEURIPS2019_9015}, TensorFlow~\cite{tensorflow2015-whitepaper}, JAX~\cite{jax2018github}, and CasADi~\cite{Andersson2019} can be applied for evaluating these Jacobians.
\end{remark}

\setlength{\textfloatsep}{1pt}
\begin{algorithm}[t]
	\caption{DiffTune }
% 	\caption{Auto-tune a controller with differentiable dynamics and controller}
	\label{algo: full algorithm using sensitivity propatation}
	\begin{algorithmic}[1]
		\REQUIRE Initial state $\tilde{\boldsymbol{x}}_0$, initial parameter $\boldsymbol{\theta}_0$, feasible set $\Theta$, horizon $N$, desired state $\{\bar{\boldsymbol{x}}_k\}_{k=0:N}$, step size $\alpha$, formulas of the Jacobians $\{\nabla_{\boldsymbol{x}} f, \nabla_{\boldsymbol{u}} f, \nabla_{\boldsymbol{x}} h, \nabla_{\boldsymbol{\theta}} h\}$, and termination condition $\mathcal{C}$.
		\ENSURE Tuned parameter $\boldsymbol{\theta}^*$
		\STATE{Initialize $\boldsymbol{\theta} \leftarrow \boldsymbol{\theta}_0$.}
		\WHILE{$\mathcal{C}$ is FALSE}
		\STATE {Set $\boldsymbol{x}_0 \leftarrow \tilde{\boldsymbol{x}}_0$ and $\partial \boldsymbol{x}_0 / \partial \boldsymbol{\theta} \leftarrow 0$.}
		\FOR{$k \gets 0$ to $N$}    
		\STATE {Obtain $\boldsymbol{x}_k$ from system and compute $\boldsymbol{u}_k$ using \eqref{eq: feedback controller}.}
		    \STATE {Update $\partial \boldsymbol{x}_{k+1} / \partial \boldsymbol{\theta}$ and $\partial \boldsymbol{u}_k / \partial \boldsymbol{\theta}$ using \eqref{eq: sensitivity propagation}.}
		    \STATE {Compute $\partial L / \partial{\boldsymbol{x}_k}$ and $\partial L / \partial{\boldsymbol{u}_k}$.}
		    \STATE {Store $\partial \boldsymbol{x}_{k+1} / \partial \boldsymbol{\theta}$, $\partial \boldsymbol{u}_k / \partial \boldsymbol{\theta}$, $\partial L / \partial{\boldsymbol{x}_k}$ and $\partial L / \partial{\boldsymbol{u}_k}$ in memory.}
        \ENDFOR
        \STATE {Compute $\nabla_{\boldsymbol{\theta}}L$ using \eqref{eq: decomposition of the target derivative to accepting sensitivity propagation} and update $\boldsymbol{\theta}$ by \eqref{eq: projected gradient descent to update the parameters}.}
        \ENDWHILE
        \RETURN{the tuned parameter $\boldsymbol{\theta}^* \leftarrow \boldsymbol{\theta}$.}
	\end{algorithmic}
\end{algorithm}

The unique aspect of sensitivity propagation is its compatibility with data from a physical system. Using such data for tuning is vital because the ultimate goal is to improve the performance of a physical system instead of its simulated counterpart. 
Despite the fidelity of the model in simulation, the physical system will have discrepancies with the model, leading to sub-optimal performance if the parameters come from simulation-based tuning. This phenomenon is part of the sim-to-real gap, which leads to degraded performance on physical systems compared to their simulated counterparts.
The sensitivity propagation, unlike the forward- or reverse-mode AD, can still be applied to compute the gradient while using data collected from the physical system.

\subsection{Auto-tuning with data from physical systems}\label{subsec: L1 for real system tuning}
The core of {\em DiffTune} is to obtain $\nabla_{\boldsymbol{\theta}}L$ from physical systems' data and then apply projected gradient descent.

However, model uncertainties and noise have to be carefully handled when using such data. Controller design usually uses the nominal model of the system, which is uncertainty- and noise-free. However, both uncertainties and noise exist in a physical system. If not dealt with, then the uncertainties and noise will contaminate the sensitivity propagation, leading to biased sensitivities and, thus, biased gradient $\nabla_{\boldsymbol{\theta}} L $, which results in inefficient parameter update. Since noise can be efficiently addressed by filtering or state estimation, our focus will be on handling the model uncertainties.

Existing methods that can compensate for the uncertainties can be applied to mitigate this issue. For example, the $\mathcal{L}_1$ adaptive control ($\mathcal{L}_1$AC) is a robust adaptive control architecture that has the advantage of decoupling estimation from control, thereby allowing for arbitrarily fast adaptation subject only to hardware limitations~\cite{hovakimyan2010L1}. It can be augmented to the controller to be tuned such that the resulting system, even though suffering from model uncertainties, behaves like a nominal system by $\mathcal{L}_1$AC's compensation for the uncertainties. 
To proceed with the illustration of how $\mathcal{L}_1$AC works, we use continuous-time dynamics to stay consistent with the notation in the majority of the $\mathcal{L}_1$AC references~\cite{hovakimyan2010L1,wang2012l1}. 
Consider the nominal system dynamics:
\begin{equation}\label{eq: nominal dynamics to explain L1}
    {\dot{\boldsymbol{x}}^\star}(t) = f({\boldsymbol{x}^\star}(t),t) + B_\text{m}({\boldsymbol{x}^\star}(t),t)\boldsymbol{u}({\boldsymbol{x}^\star}(t)), \ \boldsymbol{x}^\star(0) = \boldsymbol{x}_0,
\end{equation}
where we use $\boldsymbol{x}^\star$ to denote the nominal state, $B_\text{m} \in \mathbb{R}^{n \times m}$ to denote the control input matrix and $\boldsymbol{u}$ to denote the control input to the system.  For example, in the tuning setup, $\boldsymbol{u}$ is chosen as the baseline control 
$\boldsymbol{u}_h$ from the to-be-tuned controller $h$ in \eqref{eq: feedback controller}.

Consider the system in the presence of uncertainties:
\begin{multline}\label{eq: dynamics with uncertainty but NO L1}
    \dot{\boldsymbol{x}}(t) = f(\boldsymbol{x}(t),t) + B_\text{m}(\boldsymbol{x}(t),t)\left(\boldsymbol{u}(\boldsymbol{x}(t)) + \boldsymbol{\sigma}_\text{m}(\boldsymbol{x}(t),t)\right) \\ +B_\text{um}\boldsymbol{\sigma}_\text{um}(\boldsymbol{x}(t),t),
\end{multline}
where $\boldsymbol{\sigma}_\text{m} \in \mathbb{R}^{m}$ and $\boldsymbol{\sigma}_\text{um} \in \mathbb{R}^{n-m}$ denote the matched and unmatched uncertainties, respectively, and $\boldsymbol{x}(0) = \boldsymbol{x}_0$. The matrix $B_\text{um} \in \mathbb{R}^{n \times (n-m)}$ satisfies $B_\text{m}^\top B_\text{um}=0$ and $\text{rank}([B_\text{m} \ B_\text{um}])=n$. 
%Intuitively, the matched uncertainty $\boldsymbol{\sigma}_\text{m}$ enters the system in the same channel as the control input $\boldsymbol{u}_h$ and hence $\boldsymbol{\sigma}_\text{m}$ can be compensated for.
The uncertainty $\boldsymbol{\sigma}$, defined as $\boldsymbol{\sigma}^\top := [\boldsymbol{\sigma}_\text{m}^\top \ \boldsymbol{\sigma}_\text{um}^\top]$, poses challenges to the sensitivity propagation because the mapping $\boldsymbol{x}(t) \mapsto \boldsymbol{\sigma}(\boldsymbol{x}(t),t)$ may not be explicitly known, leaving $\partial \boldsymbol{\sigma} / \partial \boldsymbol{x}$ uncomputable in the sensitivity propagation.
%For simplicity and the clarity of exposition, we demonstrate how $\mathcal{L}_1$AC can be used for the compensation of the matched uncertainty~$\boldsymbol{\sigma}_{\text{m}}$. See \cite{li20121,xargay20101,che2012L1} for designs that tackle the unmatched uncertainty.
We consider the following control design $\boldsymbol{u}=\boldsymbol{u}_\text{h}+\boldsymbol{u}_\text{ad}$, with $\boldsymbol{u}_\text{ad}$ being the adaptive control, which results in the following system:
% We augment the control $\boldsymbol{u}_h$ with an adaptive element $\boldsymbol{u}_\text{h}+\boldsymbol{u}_\text{ad}$, with $\boldsymbol{u}_\text{ad}$ being the adaptive control, which results in the following system:
\begin{multline}\label{eq: real dynamics to explain how L1 works}
    \dot{\boldsymbol{x}}(t) = f(\boldsymbol{x}(t),t) + B_\text{um}(\boldsymbol{x}(t),t) \boldsymbol{\sigma}_\text{um}(\boldsymbol{x}(t),t) \\ + B_\text{m}(\boldsymbol{x}(t),t)(\boldsymbol{u}_h(\boldsymbol{x}(t)) +  \boldsymbol{u}_\text{ad}(t)+ \boldsymbol{\sigma}_\text{m}(\boldsymbol{x}(t),t) ).
\end{multline}
The adaptive control $\boldsymbol{u}_\text{ad}$  aims to cancel out the matched uncertainty $\boldsymbol{\sigma}_\text{m}$, i.e., $\norm{\boldsymbol{\sigma}_\text{m} + \boldsymbol{u}_\text{ad}} \approx 0$ 
% Intuitively, $\mathcal{L}_1$AC compensates for the uncertainties that are lumped from the dynamics as additive terms such that the dynamics of the real system equal to the nominal dynamics plus the additive terms 
(see \cite{hovakimyan2010L1,wu20221,pravitra2020L1,hanover2021performance} for details of how $\mathcal{L}_1$AC is implemented).
Specifically, $\mathcal{L}_1$AC estimates the uncertainty $\boldsymbol{\sigma}$ based on a state predictor and adaptation law. 
The state predictor propagates the state prediction $\hat{\boldsymbol{x}}$ based on the estimated uncertainty $\hat{\boldsymbol{\sigma}}$ and control inputs $\boldsymbol{u}_h$ and $\boldsymbol{u}_\text{ad}$, i.e.,
\begin{multline}
    \dot{\hat{\boldsymbol{x}}}(t) = f(\boldsymbol{x}(t),t) + B_\text{um}(\boldsymbol{x}(t),t) \hat{\boldsymbol{\sigma}}_\text{um}(t) + A_s(\hat{\boldsymbol{x}}(t) - {\boldsymbol{x}}(t)) \\ + B_\text{m}(\boldsymbol{x}(t),t)\left(\boldsymbol{u}_h(\boldsymbol{x}(t)) + \hat{\boldsymbol{\sigma}}_\text{m}(t) + \boldsymbol{u}_\text{ad}(t) \right),
\end{multline}
where $A_s \in \mathbb{R}^{n \times n}$ is a Hurwitz matrix, and $\hat{\boldsymbol{x}}(0) = \boldsymbol{x}_0$. The error $\hat{\boldsymbol{x}}(t) -\boldsymbol{x}(t)$ between the predicted and actual states are used to compute the estimated uncertainty, where we use the piecewise-constant adaptation law~\cite{hovakimyan2010L1}:
\begin{multline}
    \hat{\boldsymbol{\sigma}}(t) = \left[\begin{smallmatrix} \hat{\boldsymbol{\sigma}}_\text{m}(t) \\ \hat{\boldsymbol{\sigma}}_\text{um}(t)\end{smallmatrix} \right] = - [B_\text{m} \ B_\text{um}]^{-1} \text{expm}(A_sT_s - I)^{-1} \\ A_s \text{expm}(A_s T_s) (\hat{\boldsymbol{x}}(t) - \boldsymbol{x}(t)), 
\end{multline}
with $T_s$ being the sample time of $\mathcal{L}_1$AC, expm$(\cdot)$ denoting matrix exponential, and $I$ being the identity matrix.
The uncertainty's estimation error $\norm{\boldsymbol{\sigma} - \hat{\boldsymbol{\sigma}}}$ is shown to be uniformly bounded under a set of mild regularity assumptions~\cite{zhao2020adaptive,wu2023L1QuadFull}. % Lemma 3 in zhao2020adaptive and proposition 4 in wu2023L1QuadFull
Once the estimated uncertainty $\hat{\boldsymbol{\sigma}}$ is computed, the compensation $\boldsymbol{u}_\text{ad}$ is obtained by low-pass filtering $\hat{\boldsymbol{\sigma}}$, i.e., 
\begin{equation}
    \boldsymbol{u}_\text{ad}(s) = C(s)\hat{\boldsymbol{\sigma}}_\text{m}(s),
\end{equation}
with $s$ being the complex variable in the frequency domain, and $C(s)$ is the transfer function of the low-pass filter (LPF). The LPF is used here because the compensation is limited by the bandwidth of the actuator, where only the low-frequency components of $\hat{\boldsymbol{\sigma}}$ can be implemented by the actuator.
It can be shown that the residual $\norm{\boldsymbol{\sigma}_\text{m} + \boldsymbol{u}_\text{ad}}$ is bounded \cite{wang2012l1,gahlawat21a}, 
% A quick note on the boundedness of $\norm{\boldsymbol{\sigma} - \boldsymbol{u}_\text{ad}}$. It is actually achieved by adding and substracting C(s) \sigma(s), i.e., low-pass filtered uncertainty.
% ||\sigma - u_ad|| <= ||\sigma - C(s)\sigma|| + ||C(s)sigma - C(s)\hat{sigma}||
% <= ||\sigma|| ||1-C(s)|| + ||C(s)|| ||\sigma - \hat{\sigma}||
% The first term is bounded because ||sigma|| is bounded and so is ||1-C(s)||. The second term is bounded because ||C(s)|| is bounded and the estimation error ||\sigma - \hat{\sigma}|| is also bounded. 
and the error norm $\norm{{\boldsymbol{x}^\star} - \boldsymbol{x}}$ between the nominal state ${\boldsymbol{x}^\star}$ and the closed-loop state $\boldsymbol{x}$ in~\eqref{eq: real dynamics to explain how L1 works} is uniformly  bounded both in transient and steady-state~\cite{wu2023L1QuadFull,gahlawat21a}, 
which renders the uncertain system~\eqref{eq: real dynamics to explain how L1 works} behaving similar to the nominal system~\eqref{eq: nominal dynamics to explain L1}. Therefore, the sensitivity propagation remains unchanged while $\mathcal{L}_1$AC handles the uncertainties. 
We will illustrate how the $\mathcal{L}_1$AC facilitates the auto-tuning of a physical system in Sections~\ref{sec: simulation} and~\ref{sec: experiments}.

\begin{remark}
    Note that $\boldsymbol{u}_\text{ad}$ is not applied to the sensitivity propagation (only $\boldsymbol{u}_h$ is applied) because $\boldsymbol{u}_\text{ad}$ is used to cancel out the uncertainty $\boldsymbol{\sigma}_\text{m}$ to preserve the validity of the nominal dynamics~\eqref{eq: nominal dynamics to explain L1}.
\end{remark}

\subsection{Open-source DiffTune toolset}
Our toolset \texttt{DiffTuneOpenSource}~\cite{Cheng_Song_Kim} is publicly available, which facilitates users' DiffTune applications in two ways. First, it enables the automatic generation of the partial derivatives required in sensitivity propagation. In this way, a user only needs to program the dynamics and controller, eliminating the need for additional programming of the partial derivatives. Second, we provide a template that allows users to quickly set up DiffTune for custom systems and controllers. The Dubin's car and quadrotor cases used in Section~\ref{sec: simulation} are used as examples to illustrate the usage of the template.
\section{Simulation results}\label{sec: simulation}
In this section, we implement {\em DiffTune} for a Dubin's car and a quadrotor in simulations, where the controller in each case is differentiable. For all simulations, we use \texttt{ode45} to obtain the system states by integrating the continuous-time dynamics (mimicking the continuous-time process on a physical system). The states are sampled at discrete-time steps. We use the sensitivity propagation to compute $\nabla_{\boldsymbol{\theta}} L $, where the discrete-time dynamics in \eqref{eq: dynamics} are obtained by forward-Euler discretization.

We intend to answer the following questions through the simulation study: 
1. How does DiffTune compare to other auto-tuning methods? Since equipment wear is not an issue for tuning in simulations, we conduct sufficiently many trials to understand the asymptotic performance of auto-tuning methods for comparison. 2. How can the tuned parameters generalize to other unseen trajectories during tuning? 3. How does $\mathcal{L}_1$AC help tuning when the system has uncertainties? We show our main results in the following subsections while supplying the details of configurations in Appendices~A and~B.

\begin{figure}
    \centering
    \includegraphics[width = \columnwidth]{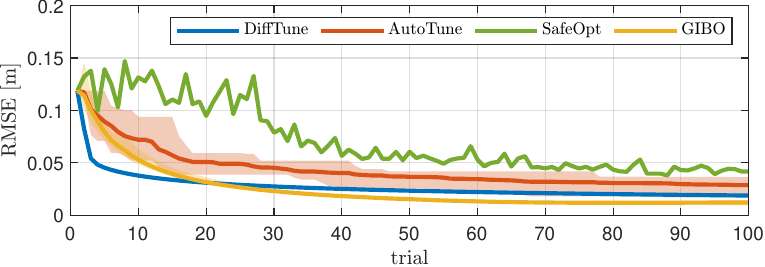}
    \caption{Comparison of auto-tuning of Dubin's car using DiffTune, AutoTune~\cite{loquercio2022autotune}, SafeOpt~\cite{berkenkamp2016safe}, and GIBO~\cite{GIBO}. The shaded areas show the range of RMSEs (min to max) achieved in a total of 10 runs of algorithms that involve stochasticity (AutoTune and GIBO).}
    \label{fig: Dubin car comparison to Autotune and SafeOpt}
\end{figure}

\subsection{Dubin's car}
\noindent\textbf{Dynamics, controller, and loss function}:
Consider the following nonlinear model:
\begin{subequations}\label{eq: nominal dynamics of Dubin's car}
\begin{align}
    \dot{x}(t) =  v(t) \cos(\psi(t)),\ \dot{y}(t) =  v(t) \sin(\psi(t)),\\
    \dot{\psi}(t) =  \omega(t), \  \dot{v}(t) =  F(t)/m, \ \dot{\omega}(t) =  M(t)/J,
\end{align}
\end{subequations}
where the state contains five scalar variables, $(x,y,\psi,v,\omega)$, which stand for horizontal position, vertical position, yaw angle, linear speed in the forward direction, and angular speed. The control actions in this model include the force $F \in \mathbb{R}$ in the forward direction of the vehicle and the moment $M \in \mathbb{R}$. The vehicle's mass and moment of inertia are known and denoted by $m$ and $J$, respectively. 
The feedback tracking controller with tunable parameter $\boldsymbol{\theta} = (k_{\boldsymbol{p}}, k_{\boldsymbol{v}}, k_{\psi}, k_{\omega})$ 
is given by
\begin{subequations}\label{eq: tracking controller for Dubin's car}
\begin{align}
    F(t) = & m ( k_{\boldsymbol{p}} \boldsymbol{e}_{\boldsymbol{p}}(t)  + k_{\boldsymbol{v}} \boldsymbol{e}_{\boldsymbol{v}}(t)+ \dot{\bar{\boldsymbol{v}}}(t))^\top \boldsymbol{q}(t), \\
    M(t) = & J ( k_{\psi}e_{\psi}(t) +  k_{\omega}e_{\omega}(t) + \dot{\bar{{\omega}}}(t)),
\end{align}
\end{subequations}
where $\bar{\cdot}$ indicates the desired value, the error terms are defined by $\boldsymbol{e}_{\boldsymbol{p}} = \bar{\boldsymbol{p}} - \boldsymbol{p}$, $\boldsymbol{e}_{\boldsymbol{v}} = \bar{\boldsymbol{v}} - \boldsymbol{v}$, $e_{\psi} = \bar{\psi} - \psi$, and $e_{\omega} = \bar{\omega} - \omega$ for $\boldsymbol{p}$ and $\boldsymbol{v}$ being the 2-dimensional vector of position and velocity, respectively, $\boldsymbol{q} = [\cos(\psi) \ \sin(\psi)]^\top$ being the heading of the vehicle, $ \bar{\boldsymbol{v}} = [\bar{v}\cos(\bar{\psi}) \ \bar{v}\sin(\bar{\phi})]^\top$ and $\dot{\bar{\boldsymbol{v}}}$ being the desired linear velocity and acceleration, respectively. 
% and $\dot{\bar{\boldsymbol{v}}} = [\dot{\bar{v}} \cos(\bar{\psi}) - \bar{v} \bar{\omega} \sin(\bar{\psi}) \ \dot{\bar{v}} \sin(\bar{\psi}) + \bar{v} \bar{\omega} \cos(\bar{\psi})]^\top$ being the vectorized desired acceleration. 
The control law \eqref{eq: tracking controller for Dubin's car} is a PD controller with proportional gains $(k_{\boldsymbol{p}}, k_{\psi})$ and derivative gains $(k_{\boldsymbol{v}}, k_{\omega})$.
If $\boldsymbol{\theta}>0$, then this controller is exponentially stable for the tracking errors $(\norm{\boldsymbol{e}_{\boldsymbol{p}}},\norm{\boldsymbol{e}_{\boldsymbol{v}}},\norm{e_{\psi}},\norm{e_{\omega}})$. We set the loss function as the RMSE of the position tracking error.
% DiffTune related results are in this directory: \Documents\GitRepo\autoDiffMatlab_DubinCar

\noindent\textbf{Comparison to other methods}: We compare DiffTune with strong baseline auto-tuning methods: AutoTune~\cite{loquercio2022autotune}, SafeOpt~\cite{berkenkamp2016safe}, and GIBO~\cite{GIBO}. \revisionnew{Note that these baseline methods are model-free probabilistic approaches that do not require knowledge of system dynamics and control.} We compare the tuning performance on a circular trajectory and assign 100 trials in each method. Other details of implementation are available in Appendix~A. The results are shown in Fig.~\ref{fig: Dubin car comparison to Autotune and SafeOpt}. 
The final tracking errors obtained by all methods are similar, and all are below 0.05 m. DiffTune achieves the fastest error reduction in the first 20 trials due to its efficient usage of the system's first-order information that can effectively guide the parameter search. 
\revision{After 20 trials, GIBO achieves the minimum error, indicating that the GP model has captured the mapping from the control parameters to the objective function.}
The performance of AutoTune and SafeOpt is inferior to that of DiffTune and GIBO, with a slower error reduction and larger errors at the end of 100 trials.
{However, GIBO and AutoTune are limited to auto-tuning in simulations because they require randomly sampled parameters for controller implementation and rollouts for performance evaluation, and then decide how to pick the next candidate parameter. This procedure leads to huge mechanical wear and tear when applied to auto-tuning of a physical system.} 
SafeOpt can produce acceptable performance by the end of the 100 trials, albeit the RMSE reduction is not smooth. The performance of SafeOpt relies on both prior knowledge (including the kernel function and its parameters and the range of feasible parameters) and the parameter space's discretization (for searching maximizers), both of which are difficult to tune (as hyperparameters in auto-tuning).

\begin{figure}
\begin{subfigure}{0.49\columnwidth}
    \centering
    \includegraphics[width = \textwidth]{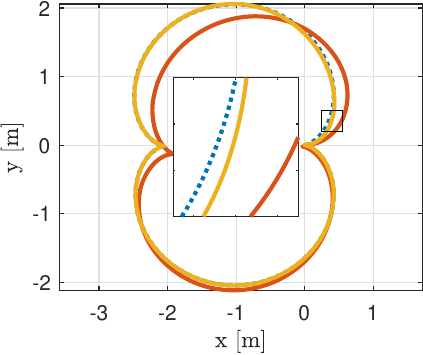}
    \caption{Peanut}
    \label{fig:peanut testing trajectory}
\end{subfigure}
\begin{subfigure}{0.49\columnwidth}
    \centering
    \includegraphics[width = \textwidth]{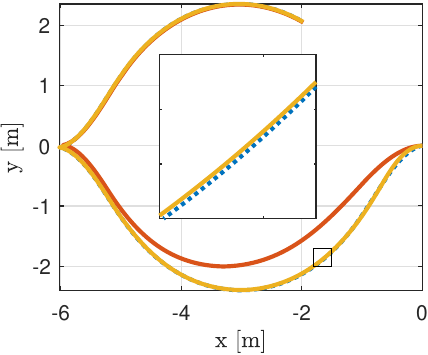}
    \caption{Lemon}
    \label{fig: Lemon testing trajecotory}
\end{subfigure}
\\
\begin{subfigure}{0.49\columnwidth}
    \centering
    \includegraphics[width = \textwidth]{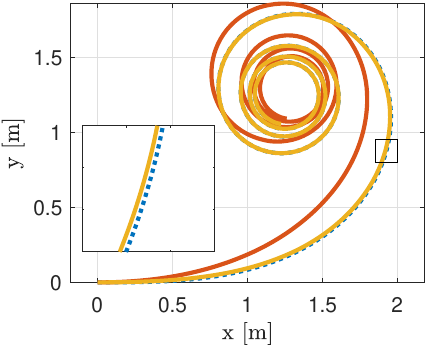}
    \caption{Spiral}
    \label{fig:Spiral testing trajectory}
\end{subfigure}
\begin{subfigure}{0.49\columnwidth}
    \centering
    \includegraphics[width = \textwidth]{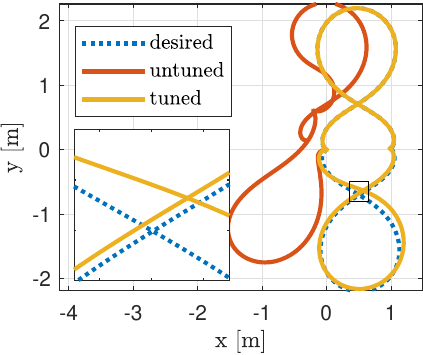}
    \caption{Twist}
    \label{fig: twist testing trajectory}
\end{subfigure}
\caption{Comparison of performance between tuned and untuned controller parameters on testing trajectories for generalization.
}
% figures are generated by dubinsCar_PDcontrol_fwdPropagation_newTraj_testing.m
    \label{fig:testing trajectories}
\end{figure}

\setlength{\tabcolsep}{5pt} % Default value: 6pt
\renewcommand{\arraystretch}{1} % Default value: 1
  \captionsetup{%size=footnotesize,
	%justification=centering, %% not needed
	skip=5pt, position = bottom}
\begin{table}[t]
	\centering
	\small
	
	\vspace{-0.25cm}
	\captionsetup{font=small}
	\caption{Testing trajectories for Dubin's car simulation and associated losses. The average loss of the tuned parameters on the tuning set is 0.36. (The units associated with linear and angular maximum speeds are m/s and rad/s, respectively.)}
	\begin{tabular}{cccrr}
		\toprule[1pt]
		\multirow{2}{*}{Trajectory} & \multicolumn{2}{c}{max speed} & \multicolumn{2}{c}{loss}  \\
		\cmidrule{2-3} \cmidrule{4-5}
		 & linear  & angular  & w/ {\em DiffTune} & w/o {\em DiffTune} \\
        \midrule
        peanut & 2 & 1.3 & \textbf{0.18} & 63.02 \\ % peanut
        lemon & 2  & 1 & \ \textbf{0.07} & 55.18 \\ % lemon
        spiral & 1 & 5 & \textbf{0.39} & 21.11\\ % spiral
        twist & 2 & 3.4 & \textbf{2.05} & 974.57\\ % twist
       	\bottomrule[1pt]
	\end{tabular}\label{tb: Dubin's car testing trajectories}
% 	\vspace{-0.3cm}
\end{table}
\normalsize

\noindent\textbf{Generalization}:
We illustrate the generalization of {\em DiffTune} in a batch tuning example. We select nine trajectories (shown in Fig.~\ref{fig:Dubins car training trajectories} in Appendix~A) as the batch tuning set. These trajectories are generated by composing constant, sinusoidal, and cosinusoidal signals for the desired linear and angular velocities. The maximum linear speed and angular speed are set to 1~m/s and 1~rad/s, respectively, to represent trajectories in one operating region. The four control parameters are all initialized at 2. The tuning proceeds by batch gradient descent on the tuning set. The controller parameters converge to $(k_{\boldsymbol{p}}, k_{\boldsymbol{v}}, k_{\psi},k_{\omega}) =  (18.83, 6.69,14.97,2.66)$. We then test the tuned parameters on four testing trajectories (unseen in the tuning set) with lemon-, twist-, {peanut-,} and spiral-shape, as shown in Fig.~\ref{fig:testing trajectories}. The tuned parameters lead to better tracking performance than the untuned ones.
The loss on the testing set is compared to the untuned parameters in Table~\ref{tb: Dubin's car testing trajectories}. It can be observed that the tuned parameters generalize well and are robust to the previously unseen trajectories.

\begin{figure}[t]
    \centering
    \includegraphics[width = \columnwidth]{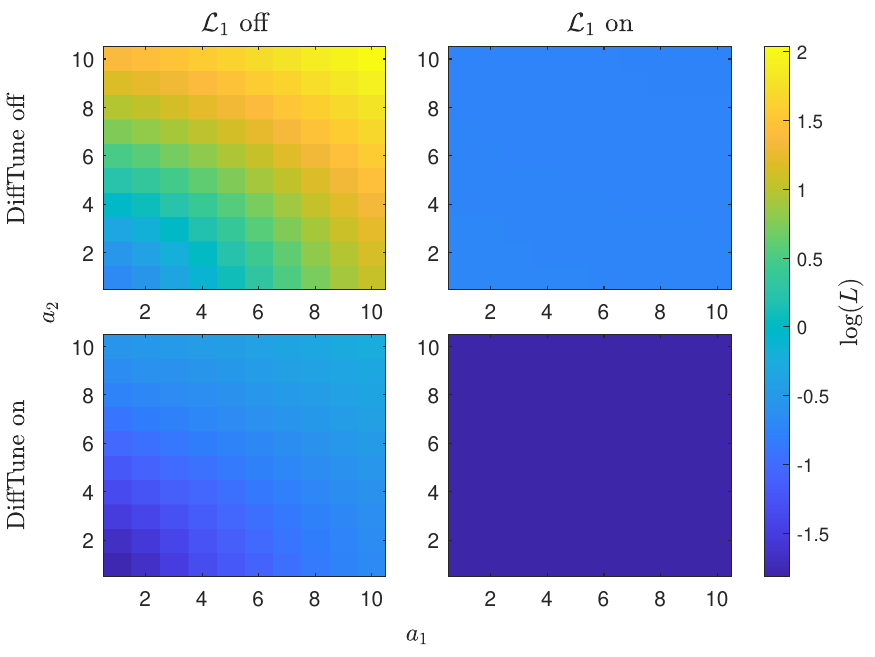}
    \vspace*{-0.6cm}
    \caption{Loss $L$ subject to uncertainties (additive force $0.1 a_1 \sin(t)$ and moment $0.1 a_2 \cos(t)$ with $a_1$ and $a_2$ taking values from 1 to 10) in the ablation study of {\em DiffTune} and $\mathcal{L}_1$AC.}
    \label{fig:loss heat map}
\end{figure}

\noindent\textbf{Handling uncertainties}:
In this simulation, we implement the $\mathcal{L}_1$AC to facilitate the  compensation for the uncertainties during tuning. For the $\mathcal{L}_1$AC, we use the piecewise-constant adaptation law and a 1st-order low-pass filter with 20 rad/s bandwidth. In this simulation, we inject additive force $0.1 a_1 \sin(t)$ and moment $0.1 a_2 \cos(t)$ to the control channels in the dynamics~\eqref{eq: nominal dynamics of Dubin's car} as uncertainties from the environment. To understand how the performance is impacted by the uncertainties, we set $(a_1, a_2)$ to a $10\times10$ grid such that $a_1$ and $a_2$ take integer values from 1 to 10, representing gradually intensified uncertainties. The four control parameters are all initialized at 10. 
We tune the controller parameters with both $\mathcal{L}_1$ on and $\mathcal{L}_1$ off, where the sensitivity propagation in both cases is based on the nominal model in~\eqref{eq: nominal dynamics of Dubin's car}. Different from the generalization test, we only tune the parameters on one trajectory (the focus is on how to reduce the impact of the uncertainties that are not considered in the nominal dynamics). The step size and termination criterion remain the same as before. 
To clearly understand the individual role of {\em DiffTune} and the $\mathcal{L}_1$AC in tuning, we conduct an ablation study. The losses are shown in Fig.~\ref{fig:loss heat map}. It can be observed that both {\em DiffTune} and $\mathcal{L}_1$AC improve the performance, and a combination of both achieves the best overall performance: the $\mathcal{L}_1$AC does so by compensating for the uncertainties, whereas {\em DiffTune} does so by driving the parameters to achieve smaller tracking error. Although the two heatmaps with $\mathcal{L}_1$ on show indistinguishable colors within each itself, the actual loss values have minor fluctuations. 

% \vspace{-0.25cm}
% \begin{figure}
%     \centering
%     \includegraphics[width = \columnwidth]{figures/parameterDistribution.pdf}
%     \vspace*{-0.6cm}
%     \caption{Tuned controller parameters subject to uncertainties (additive force $0.1 a_1 \sin(t)$ and moment $0.1 a_2 \cos(t)$).}
%     \label{fig: optimizaed parameters on the 10x10 grid}
% \end{figure}

\subsection{Quadrotor}\label{subsec: quadrotor sim}
\noindent\textbf{Dynamics, controller, and loss function}: Consider the following model on SE(3):
\begin{subequations}\label{equ:quadrotor dynamics}
\begin{align}
    & \dot{\boldsymbol{p}}=  \boldsymbol{v},\quad &&\dot{\boldsymbol{v}}=  g\boldsymbol{e}_3-\frac{f}{m}R\boldsymbol{e}_3, \label{equ:translational dynamics}\\
    & \dot{R} =  R \boldsymbol{\Omega}^{\times},\quad &&\dot{\boldsymbol{\Omega}}=  J^{-1}(\boldsymbol{M} - \boldsymbol{\Omega} \times J \boldsymbol{\Omega}) \label{equ:rotational dynamics},
\end{align}
\end{subequations}
where $\boldsymbol{p} \in \mathbb{R}^3$ and $\boldsymbol{v} \in \mathbb{R}^3$ are the position and velocity of the quadrotor, respectively, $R \in SO(3)$ is the rotation matrix describing the quadrotor's attitude, ${\boldsymbol{\Omega} \in \mathbb{R}^3}$ is the angular velocity, $g$ is the gravitational acceleration, $m$ is the vehicle mass, $J \in \mathbb{R}^{3 \times 3}$ is the moment of inertia (MoI) matrix, $f$ is the collective thrust, and $\boldsymbol{M} \in \mathbb{R}^3$ is the moment applied to the vehicle. 
The \textit{wedge} operator $\cdot^{\times}:\mathbb{R}^3 \rightarrow \mathfrak{so}(3)$ denotes the mapping to the space of skew-symmetric matrices. The control actions $f$ and $\boldsymbol{M}$ are computed using the geometric controller~\cite{lee2010geometric}. The geometric controller has a 12-dimensional parameter space, which splits into four groups of parameters: $\boldsymbol{k}_{\boldsymbol{p}}$, $\boldsymbol{k}_{\boldsymbol{v}}$, $\boldsymbol{k}_R$, $\boldsymbol{k}_{\boldsymbol{\Omega}}$ (applying to the tracking errors in position, linear velocity, attitude, and angular velocity, respectively).
Each group is a 3-dimensional vector (associated with the $x$-, $y$-, and $z$-component in each's corresponding tracking error). The initial parameters for tuning are set as $\boldsymbol{k}_{\boldsymbol{p}} = 16 \mathbb{I}$, $\boldsymbol{k}_{\boldsymbol{v}} = 5.6\mathbb{I}$, $\boldsymbol{k}_R = 8.81\mathbb{I}$, and $\boldsymbol{k}_{\boldsymbol{\Omega}} = 2.54\mathbb{I}$, for $\mathbb{I} = [1,1,1]^\top$. The feasible sets of controller parameters are set as $\boldsymbol{k}_{\boldsymbol{p}} \in [15,24]$,
$\boldsymbol{k}_{\boldsymbol{v}} \in [4, 16] $, 
$\boldsymbol{k}_{\boldsymbol{R}} \in [8, 12] $, and
$\boldsymbol{k}_{\boldsymbol{\Omega}} \in [0.1, 3]$.
We set the loss function as the squared norm of the position tracking error, summed over a horizon of 10~s. 
We add zero-mean Gaussian noise to the position, linear velocity, and angular velocity (with standard deviation 0.1~m, 0.1~m/s, 1e-3~rad/s, respectively). 

\noindent\textbf{Comparison to other methods}: We compare DiffTune with strong baselines AutoTune~\cite{loquercio2022autotune}, SafeOpt-PSO~\cite{duivenvoorden2017constrained}, and GIBO~\cite{GIBO}. \revisionnew{Note that these baseline methods are model-free probabilistic approaches that do not require knowledge of system dynamics and control.} The middle is a variant of the SafeOpt~\cite{berkenkamp2016safe}, which applies Particle Swarm Optimization (PSO) to enable adaptive discretization of the parameter space. The original SafeOpt is not applicable because it requires fine discretization of the parameter space to search for the maximizer, which suffers from the curse of dimensionality. Specifically, auto-tuning of the geometric controller requires at least $12^M$ discretization points if each parameter admits at least $M$ discretization points.

\begin{figure}[h]
    % \centering
    % \includegraphics[width = \columnwidth]{figures/loss_curve_lr.png}
    % \vspace*{-0.5cm}
    \begin{subfigure}{\columnwidth}
        \includegraphics[width = \columnwidth]{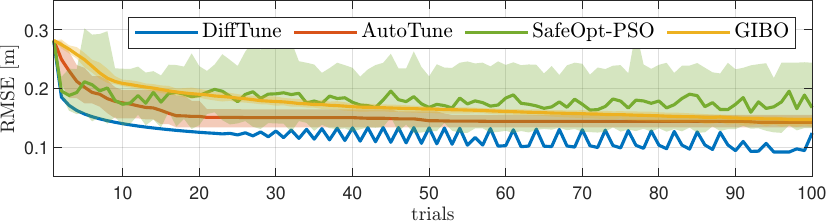}
        \caption{2D circle}
        \label{fig: 2D circle simulation comparison}
    \end{subfigure}
    \begin{subfigure}{\columnwidth}
        \includegraphics[width = \columnwidth]{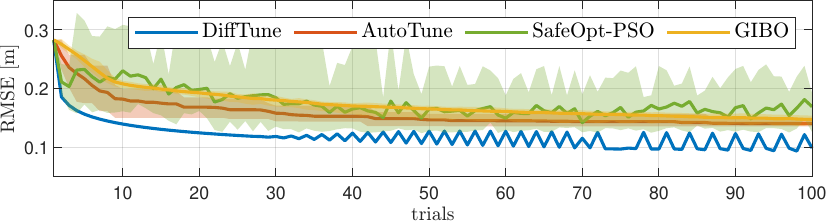}
        \caption{3D circle}
        \label{fig: 3D circle simulation comparison}
    \end{subfigure}
    \begin{subfigure}{\columnwidth}
        \includegraphics[width = \columnwidth]{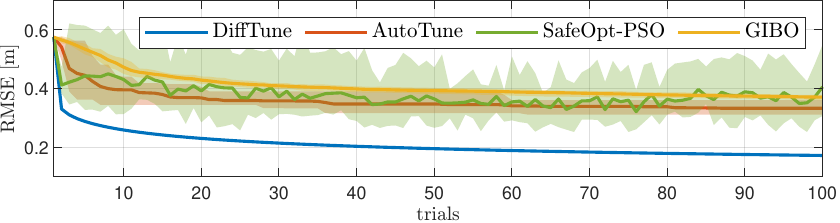}
        \caption{3D figure 8}
        \label{fig: 3D figure 8 simulation comparison}
    \end{subfigure}
    \caption{Comparison of tuning the controller of a quadrotor using DiffTune (proposed), AutoTune~\cite{loquercio2022autotune}, SafeOpt-PSO~\cite{duivenvoorden2017constrained}, and GIBO~\cite{GIBO}. The shaded areas show the range of RMSEs (min to max) achieved in a total of 10 runs of the algorithms that involve stochasticity (AutoTune, SafeOpt-PSO, and GIBO).}
  \label{fig:quadrotor_sim difftune autotune safeopt}
\end{figure}

The detailed settings of AutoTune and SafeOpt in this example are shown in Appendix~B.
We compare the three auto-tuning methods on three trajectories, where 100 trials are performed for each method on each trajectory. The results are shown in Fig.~\ref{fig:quadrotor_sim difftune autotune safeopt}, where DiffTune achieves the minimum tracking RMSE with the best efficiency. Note that the auto-tuning of the quadrotor is more complicated than that of Dubin's car due to the former's higher dimensional parameter space and stronger nonlinearities in dynamics and control. In the auto-tuning on the 2D/3D circular trajectories, the RMSEs show oscillation near the end of the tuning trials, indicating the learning rate might be too large when the loss is close to a (local) minimum. 
\revision{In terms of the final RMSE, AutoTune, SafeOpt-PSO, and GIBO demonstrate similar performance, all inferior to that of DiffTune. AutoTune and GIBO have a smoother RMSE reduction than SafeOpt-PSO. Moreover, the three baseline methods are less favorable for practical usage since they demand more hyperparameters to be tuned (e.g., the variance of the transition model for each parameter in AutoTune; kernel functions, lower/upper bound of each parameter, safety thresholds, and swarm size for SafeOpt-PSO; kernel functions, local search bounds, learning rate, and number of queries for GIBO) than the number of controller parameters for auto-tuning.} 
%For example, a large swarm size allows for a broader search space. However, this advantage comes with increased computational complexity. Also, the bounds of the parameter need to be set carefully. Large bounds can increase the possibility of finding better solutions but reduce the efficiency of the search algorithm.
In contrast, DiffTune only requires tuning the learning rate as the sole hyperparameter, yet still delivering the best outcome.

\noindent\textbf{Handling uncertainties}: In this simulation, we consider the uncertainty caused by the imprecise knowledge of the moment of inertia (MoI) $J$. We set the vehicle's true MoI as $\beta J$ for $\beta$ from 0.5 to 4 and use $J$ in the controller design as our best knowledge of the system. The scaled MoI can be treated as an unknown control input gain (see \eqref{equ:rotational dynamics}), leading to decreased ($\beta>1$) or increased ($\beta < 1$) moment in reality compared to the commanded moment by the geometric controller. However, the uncertainty caused by the perturbed MoI can be well handled by $\mathcal{L}_1$AC, which is adopted in the simulation (formulation detailed in \cite{wu20221}).
We conduct an ablation study to understand the roles of {\em DiffTune} and $\mathcal{L}_1$AC by comparing the root-mean-square error (RMSE) of position tracking, as shown in Table~\ref{tb: quadrotor RMSE table} (where the tuning is conducted over a 3D figure 8 trajectory in 100 trials with a learning rate of $\alpha=0.005$). It can be seen that tuning and $\mathcal{L}_1$ can individually reduce the tracking RMSE. The best performance is achieved when DiffTune and $\mathcal{L}_1$AC are applied jointly.

\setlength{\tabcolsep}{5pt} % Default value: 6pt
\renewcommand{\arraystretch}{1} % Default value: 1
  \captionsetup{%size=footnotesize,
	%justification=centering, %% not needed
	skip=5pt, position = bottom}
\begin{table}[h]
	\centering
	\small
	
	\vspace{-0.2cm}
	\captionsetup{font=small}
	\caption{Tracking RMSE [cm] with MoI perturbation $\beta$ ranging from 0.5 to 4 in the ablation study of DiffTune and $\mathcal{L}_1$AC. (DT stands for ``{\em DiffTune}.'')}
	\begin{tabular}{lcccccccc}
		\toprule[1pt]
		\multicolumn{1}{c}{$\beta$} & 0.5 & 1 & 1.5 & 2 & 2.5 & 3 & 3.5 & 4
		\\
        % \midrule
        % T \ $\mathcal{L}_1$ &  & &  &  &  &  &  & 
        % \\
		\midrule
		% tuning on L1 on
		DT \ $\mathcal{L}_1$ & \textbf{30.5} & \textbf{30.1} & \textbf{30.8} & \textbf{32.9} & \textbf{34.3} & \textbf{35.4} & \textbf{36.5} & \textbf{37.9}\\
		% tuning on L1 off
		DT \ \sout{$\mathcal{L}_1$} & 30.8 & \textbf{30.1} &  31.0 & 33.9 & 37.8 & 39.5 & 40.5 & 41.1\\
		% tuning off L1 on
		\sout{DT} \ $\mathcal{L}_1$ & 57.9 & 57.5 &  58.0 & 59.5 & 62.2 & 66.0 & 70.9 & 76.8\\
		% tuning off L1 off
		\sout{DT} \ \sout{$\mathcal{L}_1$} & 57.8 & 57.6 & 58.9 & 62.2 & 67.3 & 74.1 & 82.2 & 90.9\\
       	\bottomrule[1pt]
	\end{tabular}\label{tb: quadrotor RMSE table}
\end{table}

\subsection{Discussion}
The advantage of DiffTune is its efficient usage of the first-order information (gradient $\nabla_{\boldsymbol{\theta}}L$) of the target system. \revisionnew{Compared with DiffTune, the baseline auto-tuning methods (AutoTune, SafeOpt/SafeOpt-PSO, and GIBO) have the advantage of requiring less prior information (e.g., explicit formulas for dynamics~\eqref{eq: dynamics} and controller~\eqref{eq: feedback controller}) than DiffTune.}
% tuning without requiring the knowledge of the model (either when the precise model is difficult to establish or when the model/performance metric makes the parameter optimization problem~\eqref{prob: controller tuning as a parameter optimization} difficult to solve). 
All baseline methods use probabilistic approaches (Metropolis-Hastings algorithm or Bayesian optimization) to explore candidates of parameters and iteratively improve the performance based on observed input-output (i.e., parameters-performance) pairs. However, when the knowledge of the system is perceived through such ``sampling'' procedures, sufficiently many trials are needed to gain enough information and infer the optimal parameter choice, and the number of trials scales badly with the dimension of the parameter space. However, in practice, many physical systems have models obtained by physics or first principles, which can provide sufficiently useful first-order information to guide parameter searches. Such information will significantly reduce the number of trials in auto-tuning compared to when one uses ``sampling'' to obtain this information, which is clear in the comparison shown in Figs.~\ref{fig: Dubin car comparison to Autotune and SafeOpt} and~\ref{fig:quadrotor_sim difftune autotune safeopt}.
\revision{For BO-based approaches (SafeOpt/SafeOpt-PSO and GIBO), the assumption that the objective function is a sample from a known GP prior may not fit the auto-tuning scenario, especially when the system dynamics or controller hold strong nonlinearities. This conclusion is drawn from the observation of BO-based approaches' \textit{inferior} performance to DiffTune for quadrotor auto-tuning in Fig.~\ref{fig:quadrotor_sim difftune autotune safeopt}, in contrast to the \textit{similar} performance obtained by DiffTune and BO-based approaches for the auto-tuning of Dubin's car (whose dynamics and controller exhibit less nonlinearities than those of the quadrotor) in Fig.~\ref{fig: Dubin car comparison to Autotune and SafeOpt}.
}
We will illustrate the efficiency of auto-tuning using the first-order information with the experimental results in Section~\ref{sec: experiments} next.

\section{Experiment results}\label{sec: experiments}
We validate and evaluate DiffTune on a quadrotor in experiments, through which we would like to answer the following four questions: 1. How is the performance improvement using DiffTune with only limited tuning budgets (e.g., 10 trials)? 2. How do the tuned parameters generalize to trajectories that are unseen during tuning? 3. What are the individual role of DiffTune and $\mathcal{L}_1$AC in terms of performance improvement? 
\revision{4. How is the real-flight performance compared between parameters auto-tuned with experimental data and those auto-tuned with simulation data?} % \textcolor{red}{Whether the tuning performance degradation due to the sim-to-real gap is acceptable/significant, i.e., how the performance of control parameters obtained utilizing simulation data and experimental data compare with each other on a real experiment platform?}
% \revision{4. How is the performance comparison between simulation-based auto-tuning and DiffTune on a real platform?} \textcolor{red}{explain the term difference between simulation-based and experimental data-based auto-tuning. use textit to highlight the difference of compared approaches. they only differ in the type of data: simulated vs real. use foot note. Keep the question short and clear. Lin suggests to use subsection E title because it's clear}

\subsection{Experiment setup}We use the same dynamics and controller as used in Section~\ref{subsec: quadrotor sim}. The controller's initial parameters are shown in Table~\ref{tb: parameters summary for quadrotor experiment}. 
We only permit 10 tuning trials as the budget to limit the time and mechanical wear of the tuning. 
\revision{The loss function is chosen as the sum of the translational and rotational tracking errors to penalize undesirable tracking performance in these two perspectives, i.e., $L = \lVert \boldsymbol{p}-\bar{\boldsymbol{p}} \rVert^2 + \text{tr}(I - \bar{R}^\top R)  /2$.}
A horizon of 7 s is used to collect the data and perform sensitivity propagation. The data contains the full state and control actions sampled at 400 Hz (by the design of Ardupilot), where the state is obtained via the original EKF developed by Ardupilot, with Vicon providing only position and yaw measurements of the quadrotor. 
The data are logged onboard and downloaded to a laptop to compute the new controller parameter $\boldsymbol{\theta}$.
We use learning rate $\alpha = 0.1$ together with gradient clipping such that the parameters in the next trial $\boldsymbol{\theta}_{j+1}$ will always fall within 10\% of the current parameters $\boldsymbol{\theta}_{j}$, i.e., $\boldsymbol{\theta}_{j+1} \in [0.9\boldsymbol{\theta}_{j},1.1\boldsymbol{\theta}_{j}]$. Such a saturation scheme is used to (i) prevent the parameters from turning negative when the gradient is large and (ii) enforce a ``trust region'' around the current parameters to avoid overly large parameter changes. The laptop has an Intel i9-8950HK CPU, and the run time for sensitivity propagation to update the sensitivity states in one iteration (from $k$ to $k+1$) is $91 \pm 13 \mu$s (in MATLAB).

\subsection{Run DiffTune on three trajectories}\label{subsec: tuning on three trajectories} 

\begin{figure*}
    \begin{subfigure}{0.33\textwidth}
    \centering
    \includegraphics[width = \textwidth]{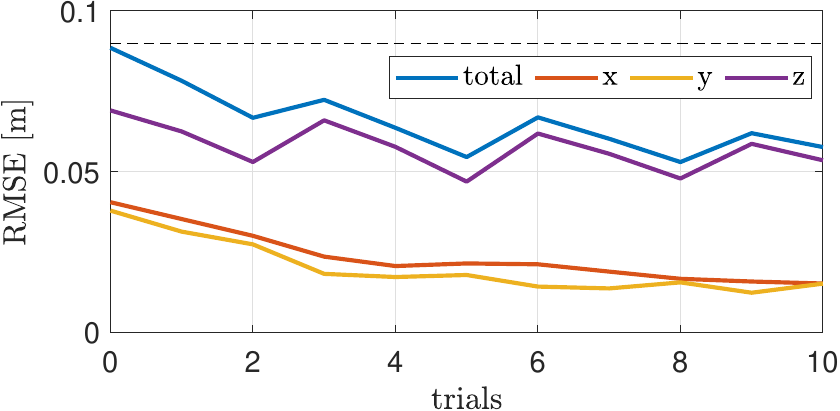}
    \caption{1 m/s}
    \label{fig: quad exp RMSE 1mps}
\end{subfigure}
\begin{subfigure}{0.33\textwidth}
    \centering
    \includegraphics[width = \textwidth]{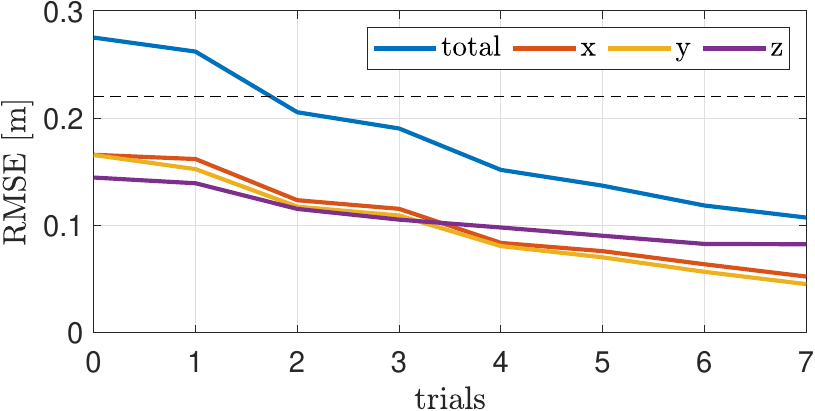}
    \caption{2 m/s}
    \label{fig:quad exp RMSE 2mps}
\end{subfigure}
\begin{subfigure}{0.33\textwidth}
    \centering
    \includegraphics[width = \textwidth]{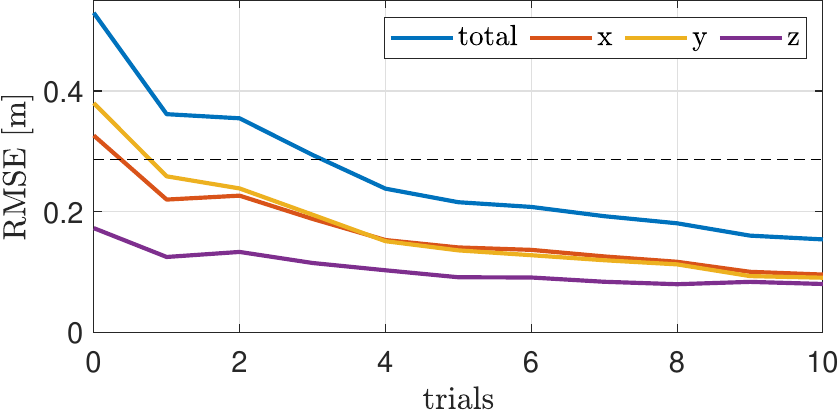}
    \caption{3 m/s}
    \label{fig:quad exp RMSE 3mps}
\end{subfigure}
\caption{Tracking error (RMSE) of the tuning on three circular trajectories. The dashed line shows the RMSE achieved by hand-tuning.}
\label{fig: RMSE exp}
\end{figure*}

\begin{figure*}
    \begin{subfigure}{0.33\textwidth}
    \centering
    \includegraphics[width = \textwidth]{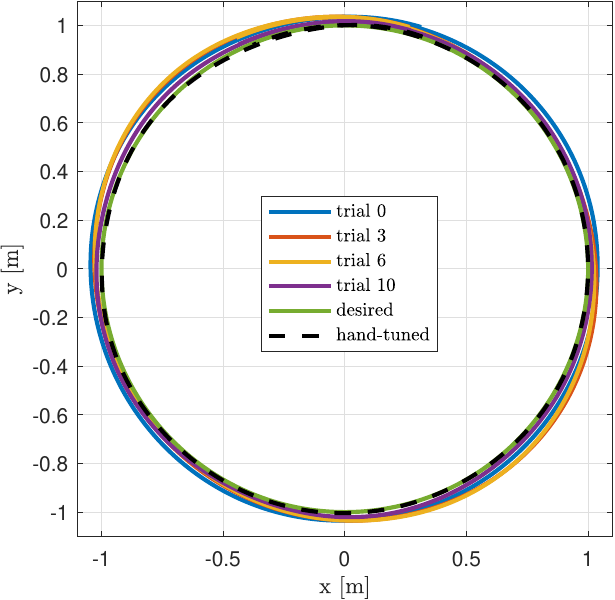}
    \caption{1 m/s}
    \label{fig: quad exp traj 1mps}
\end{subfigure}
\begin{subfigure}{0.33\textwidth}
    \centering
    \includegraphics[width = \textwidth]{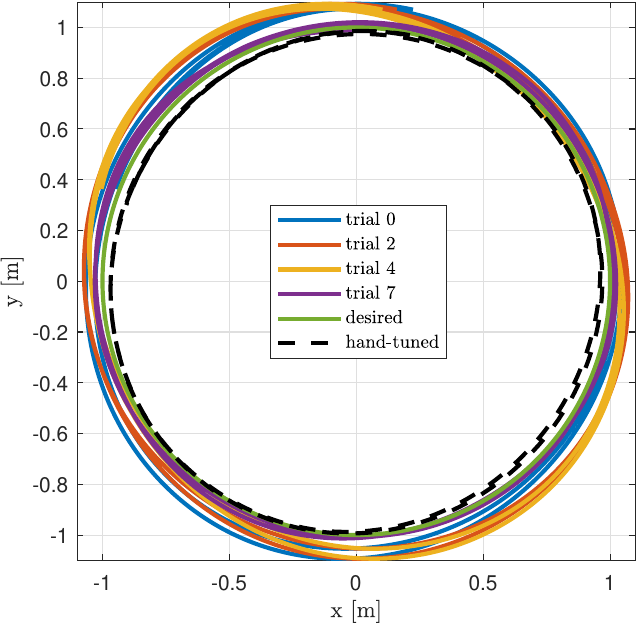}
    \caption{2 m/s}
    \label{fig:quad exp traj 2mps}
\end{subfigure}
\begin{subfigure}{0.33\textwidth}
    \centering
    \includegraphics[width = \textwidth]{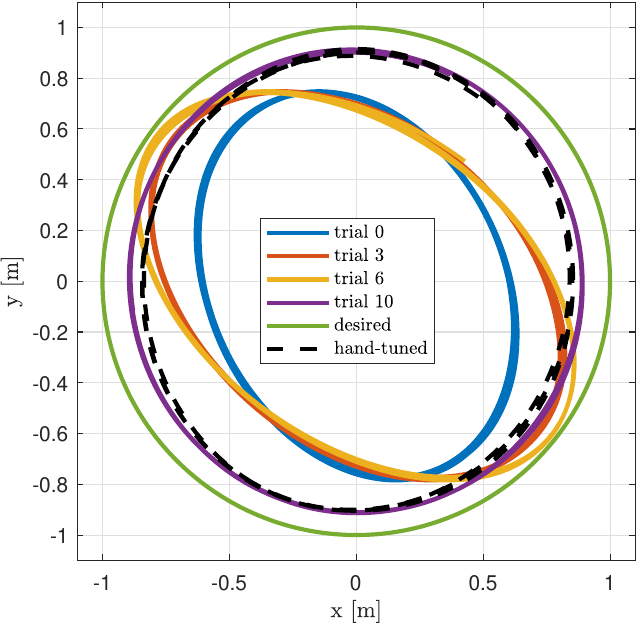}
    \caption{3 m/s}
    \label{fig:quad exp traj 3mps}
\end{subfigure}
\caption{Horizontal trajectories of the quadrotor during tuning on three circular trajectories with different speeds.}
\label{fig: trajectory exp}
\end{figure*}

\begin{figure*}
    \begin{subfigure}{0.33\textwidth}
    \centering
    \includegraphics[width = \textwidth]{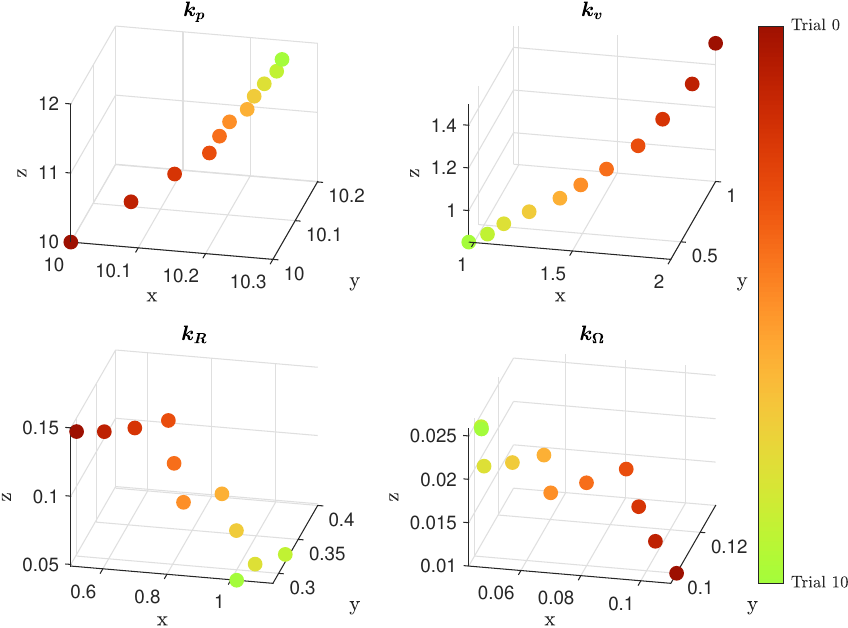}
    \caption{1 m/s}
    \label{fig: quad exp params 1mps}
\end{subfigure}
\begin{subfigure}{0.33\textwidth}
    \centering
    \includegraphics[width = \textwidth]{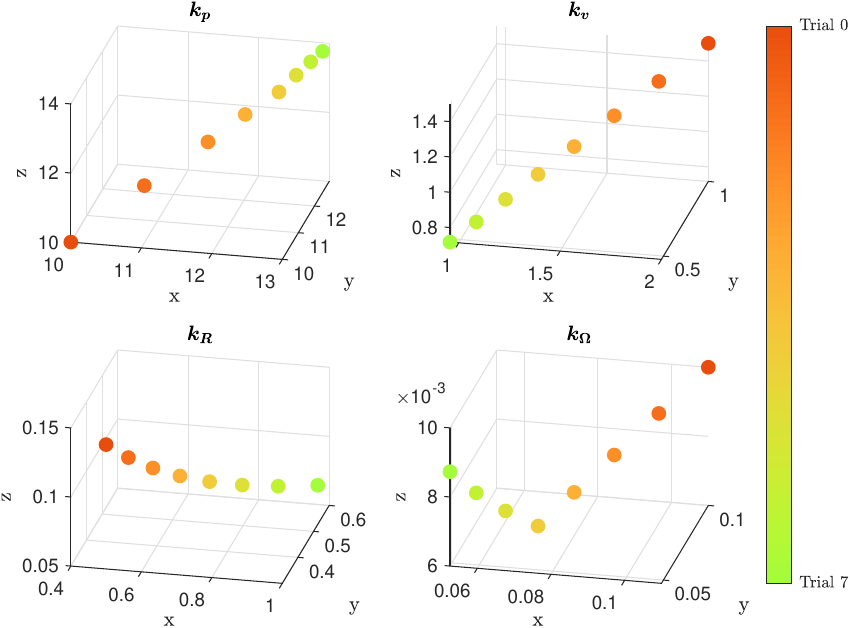}
    \caption{2 m/s}
    \label{fig:quad exp params 2mps}
\end{subfigure}
\begin{subfigure}{0.33\textwidth}
    \centering
    \includegraphics[width = \textwidth]{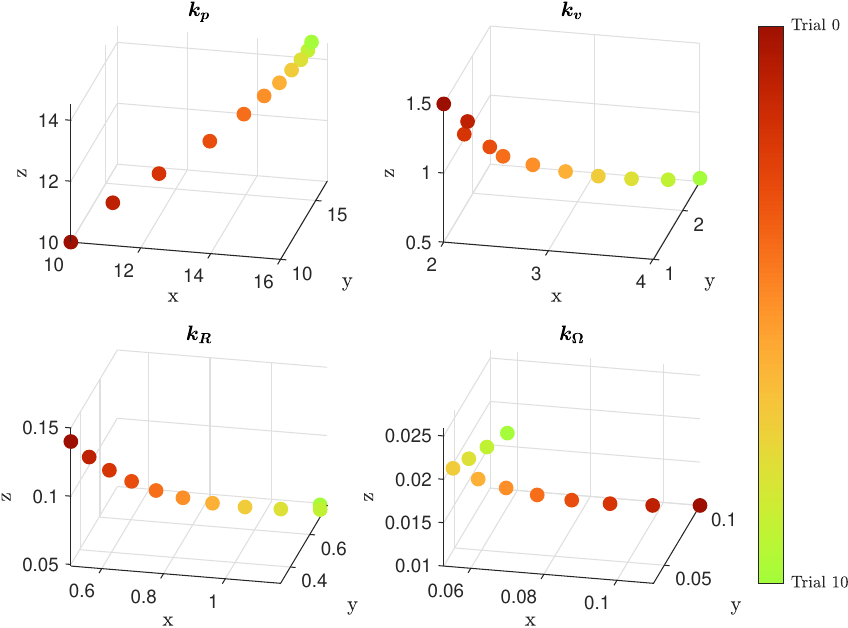}
    \caption{3 m/s}
    \label{fig:quad exp params 3mps}
\end{subfigure}
\caption{History of parameters during the tuning on three circular trajectories with different speeds.}
\label{fig: params exp}
\end{figure*}

We use a circular trajectory for tuning, where the speeds are set to $[1,2,3]$ m/s for a spectrum of agility from slow to aggressive. The controller is tuned individually for these three speeds\footnote{The video recordings of the 0th, 3rd, 6th, and 10th trials while tuning for the 3 m/s circular trajectory are available in the supplementary material, in which one can see the performance improvement through the trials.}, and we denote the final parameters by P1, P2, and P3 (associated with the speeds of 1, 2, and 3 m/s, respectively). The parameters P2 are obtained in seven trials because the quadrotor experiences oscillations after the seventh trial, and we decided to use the parameters tuned at the last non-oscillating trial. The reduction of the tracking RMSE is shown in Fig.~\ref{fig: RMSE exp}. Comparing the tracking performance at the last trial to the initial trial, the RMSE has achieved 1.5x, 2.5x, and 3.5x reduction on the 1, 2, and 3 m/s circular trajectories, respectively. Furthermore, all the tuned parameters achieve lower tracking RMSEs than the hand-tuned parameters.
While the reduction in tracking RMSE is monotone in the 2 and 3 m/s cases, for the case of 1 m/s, a minor fluctuation is superposed on the monotone reduction of the tracking RMSE. This phenomenon happens since the dominant $z$-axis RMSE fluctuates, which is caused by the large learning rate for $z$-axis tracking when the parameters are close to the (local) minimum of $z$-axis error (observe the $z$-axis RMSE reduces only from 6.8 to 5.5 cm). For the $x$- and $y$-axis tracking RMSE, we observe a monotone reduction in all three speeds. Notably, we observe the sensitivity of the angular velocity gains $\boldsymbol{k}_{\boldsymbol{\Omega}_x}$ and $\boldsymbol{k}_{\boldsymbol{\Omega}_y}$ are larger than the other parameters (by at least a magnitude). In other words, the partial derivatives $\partial L / \partial \boldsymbol{k}_{\boldsymbol{\Omega}_x}$ and $\partial L / \partial \boldsymbol{k}_{\boldsymbol{\Omega}_y}$ are large, which leads to significant changes in the gains $\boldsymbol{k}_{\boldsymbol{\Omega}_x}$ and $\boldsymbol{k}_{\boldsymbol{\Omega}_y}$. This reduction leads to a more agile response in rotational tracking on the roll and pitch commands, which efficiently improves the position tracking performance on the $x$- and $y$-axis. 
The trajectories on the horizontal plane through the tuning trials are shown in Fig.~\ref{fig: trajectory exp}. For the 3 m/s circular tuning trajectory, we show the stacked images during the flight in Fig.~\ref{fig: exp stacked images 3mps}. It is clear that the quadrotor's tracking of the circular trajectory becomes better as more trials are conducted.

The evolution of the parameters in the tuning trials is shown in Fig.~\ref{fig: params exp}. Overall, the parameters tend to converge following one direction, except for the derivative gain $\boldsymbol{k}_{\boldsymbol{\Omega}}$ for angular tracking. Specifically, near the end of the tuning, $\boldsymbol{k}_{\boldsymbol{\Omega}}$ shows a change of evolution direction, which indicates that the mapping from these parameters to the loss function is likely to lie on a nonlinear manifold. Such a nonlinear manifold is difficult for a human to perceive and understand in hand-tuning unless sufficiently many, possibly pessimistically unrealistically many, trials are provided, which leads to the challenges in tuning a nonlinear controller by hand. Furthermore, another challenge of hand tuning is that one may alter one or (at most) two parameters in each trial since humans essentially perform coordinate-wise finite differences via trial and error to tune the controller. These two factors combined result in inefficient tuning by hand, especially when the dimension of parameter space is high (e.g., 12 for the geometric control~\cite{lee2010geometric}). We display the tuned parameters in Table~\ref{tb: parameters summary for quadrotor experiment} in Appendix~C, along with the initial parameters and hand-tuned parameters for comparison.

\subsection{Generalization} 

We conducted experiments to test the generalization capability of the tuned parameters in Section~\ref{subsec: tuning on three trajectories}. The testing set contains circular, 3D figure 8, and vertical figure 8 trajectories, with their coordinates shown in Table~\ref{tb: trajectories used for generalizability} and shapes shown in Fig.~\ref{fig: trajectory shape for generalizability} (in the Appendix~C). The latter two trajectories are considered here for their wide range of speed and acceleration (see Table~\ref{tb: experiment generalizability test result}), which is in contrast to those static values of the circular trajectories. We test the three groups of tuned parameters P1, P2, and P3 on circular trajectories with 1, 2, and 3 m/s speed, respectively. The tracking performance of these tuned parameters on the testing trajectories is shown in Table~\ref{tb: experiment generalizability test result}, in which we also include the baseline of hand-tuned parameters.

\setlength{\tabcolsep}{5pt} % Default value: 6pt
\renewcommand{\arraystretch}{1} % Default value: 1
  \captionsetup{%size=footnotesize,
	%justification=centering, %% not needed
	skip=5pt, position = bottom}
\begin{table}[h]
	\centering
	\small
	
	\vspace{-0.2cm}
	\captionsetup{font=small}
	\caption{Trajectories used in the generalizability study. The trajectories are parameterized by $v$, and their $xyz$-coordinates are shown in the ``Function'' column ($t$ is the time argument).}
	\begin{tabular}{lll}
		\toprule[1pt]
  % speed
		Trajectory & Notation & Function 
		\\
		\midrule
		% circular
                 & & $x(t) = \sin(vt)$  \\
		Circular &$C(v)$  & $y(t) = -\cos(vt)$  \\
                 & & $z(t) = -1$  \\
                 \midrule
        % 3D figure 8
                 & & $x(t) = 1.5\sin(vt/1.75)$  \\
		3D figure 8 & $DF(v)$ & $y(t) = \sin(2vt/1.75)$  \\
                 & & $z(t) = -1-0.3\sin(0.5vt/1.75)$ \\
                 \midrule
        % vertical figure 8
                 & & $x(t) = 1.5\sin(vt/1.25)$  \\
		Vertical figure 8 & $VF(v)$ & $y(t) = \sin(vt/1.25)$  \\
               &  & $z(t) = -1-0.3\sin(2vt/1.25)$  \\
       	\bottomrule[1pt]
	\end{tabular}\label{tb: trajectories used for generalizability}
	% \vspace{-0.5cm}
\end{table}
\normalsize

\setlength{\tabcolsep}{5pt} % Default value: 6pt
\renewcommand{\arraystretch}{1} % Default value: 1
  \captionsetup{%size=footnotesize,
	%justification=centering, %% not needed
	skip=5pt, position = bottom}
\begin{table}[t]
	\centering
	\small
	
	% \vspace{-0.2cm}
	\captionsetup{font=small}
	\caption{Tracking RMSE [cm] in the generalizability test. The star indicates minor oscillations on the pitch angle of the quadrotor. ``HT'' stands for hand-tuned.}
	\begin{tabular}{lcc|ccc|c}
		\toprule[1pt]
        Traj. & Spd. & Acc. & P1 & P2 & P3 & HT \\
         & [m/s] & [m/s${}^2$] & [cm] & [cm] & [cm] & [cm]\\
        \midrule
        $C(1)$ & 1 & 1 & \textbf{5.7} & \textbf{5.7} & 5.5* & 9.0\\
        $C(2)$ & 2 & 4 & 27.2 & \textbf{10.7} & 16.6 & 18.0\\
        $C(3)$ & 3 & 9 & 72.8 & 31.7 & \textbf{15.4} & 28.6\\
        \midrule
        $DF(1)$ & 0.59--1.43 &  0.00--1.35 & 5.5 & \textbf{4.9} & 4.8* & 5.3 \\
        $DF(2)$ & 1.17--2.86 &  0.00--5.41 & 10.6 & 8.8 & \textbf{8.3} & 9.2 \\
        \midrule
        $VF(1)$ & 0.48--1.52 &  0.00--1.20 & 12.5 & 10.5 & 9.7* & \textbf{9.6} \\
        $VF(2)$ & 0.96--3.04 &  0.00--4.81 & 17.8 & 13.0 & \textbf{9.8} & 12.7 \\
       	\bottomrule[1pt]
	\end{tabular}
 \label{tb: experiment generalizability test result}
	% \vspace{-0.5cm}
\end{table}
\normalsize

When tested on the circular trajectories, the tuned parameters perform the best on the speed that they were tuned for, i.e., P$n$ performs the best on trajectory $C(n)$ for $n \in \{1,2,3\}$. The same phenomenon has been observed in~\cite{loquercio2022autotune} (although the controller therein is different from the one used here), where parameters perform the best over the trajectories that they are auto-tuned on. This behavior is similar to an overfitted NN in machine learning. In our case, this type of ``overfitting'' is expected since the proportional-derivative structure of the geometric controller determines that there does not exist a group of parameters that work well for all conditions (e.g., the aggressive trajectory $C(3)$ demands a distinct parameter combination than for the slow trajectory $C(1)$). However, the parameters can still generalize to the 3D figure 8 and vertical figure 8 trajectories that are not used for tuning. Specifically, P2 and P3, which are tuned for increasingly aggressive maneuvers with fast-changing directions of velocity and acceleration, generalize to the two variants of the figure 8 trajectory that demands fast-changing speed and acceleration. P3 demonstrates better agility, which shows the best performance on $DF(2)$ and $VF(2)$. However, P3 is overly agile for the speed and acceleration in $DF(1)$ and $VF(1)$, which leads to minimum RMSE compared to P1 and P2, albeit with minor oscillations on the pitch angle.

\subsection{Ablation Study}\label{subsec: ablation with L1}
We conduct an ablation study of how much contribution DiffTune and $\mathcal{L}_1$AC provide to performance improvement. We repeat the tuning in Section~\ref{subsec: tuning on three trajectories} but with $\mathcal{L}_1$AC in the loop, which results in different sets of parameters for the three trajectories denoted by Pn-$\mathcal{L}_1$ for $n \in \{1,2,3\}$ and shown in Table~\ref{tb: parameters summary for quadrotor experiment} in Appendix~C. Our implementation follows~\cite{wu2023mathcal}. Unlike the simulations in Section~\ref{subsec: quadrotor sim} where we deliberately introduce uncertainties in MoI, in the experiment, we do not introduce uncertainties. The quadrotor naturally has uncertainties existing in the system (e.g., varying battery voltage in flight and mismatch between the actual MoI and estimated MoI through CAD computation), in which case $\mathcal{L}_1$AC can help compensate for these uncertainties and thus improve the tracking performance. The results are shown in Table~\ref{tb: ablation study with L1}. Here, ``DiffTune off'' shows the performance of the initial parameters (no tuning occurs); ``DiffTune on'' shows the tracking performance in the final tuning trial. ``$\mathcal{L}_1$ on/off'' indicate whether $\mathcal{L}_1$AC is used or not during the tuning trials.  
We conclude that 1. DiffTune alone can improve the tracking performance, albeit the system has uncertainties and measurement noise. 2. When $\mathcal{L}_1$AC is combined with DiffTune, $\mathcal{L}_1$AC helps improve the performance in two ways: (i) compensating for the uncertainties so that the uncertainties' degradation to system performance is mitigated; and (ii) DiffTune can proceed with less biased gradient thanks to the uncertainties being ``canceled out'' by $\mathcal{L}_1$AC, which leads to more efficient tuning.
Intuitively, DiffTune raises the performance's upper limit (in an ideal case subject to no uncertainty), whereas $\mathcal{L}_1$AC keeps the actual performance (in a realistic case subject to uncertainties) close to the upper limit. When the two are used together, the best performance is achieved. 

\setlength{\tabcolsep}{5pt} % Default value: 6pt
\renewcommand{\arraystretch}{1} % Default value: 1
  \captionsetup{%size=footnotesize,
	%justification=centering, %% not needed
	skip=5pt, position = bottom}
\begin{table}[h]
	\centering
	\small
	
	\vspace{-0.4cm}
	\captionsetup{font=small}
	\caption{Tracking RMSE [cm] in ablation study with DiffTune and $\mathcal{L}_1$AC.}
	\begin{tabular}{cc|rr|rr|rr}
		\toprule[1pt]
  % speed
		\multicolumn{2}{c|}{Speed [m/s]} & \multicolumn{2}{c|}{1} & \multicolumn{2}{c|}{2} & \multicolumn{2}{c}{3} 
		\\
		\midrule
		% L1 indication
		\multicolumn{2}{c|}{$\mathcal{L}_1$AC} & off & on & off & on & off & on \\
  \midrule
		% DiffTune off
		\multirow{ 2}{*}{DiffTune} & off  & 8.9 & 7.5 & 27.5 & 25.1 & 61.8 & 46.6\\
		% DiffTune on
		 & on  & 5.7 & \textbf{3.0} & 10.7 & \textbf{6.9} & 17.7 & \textbf{16.2} \\
       	\bottomrule[1pt]
	\end{tabular}\label{tb: ablation study with L1}
	% \vspace{-0.5cm}
\end{table}
\normalsize

\subsection{\revision{Comparison of Parameters Auto-tuned in Experiments with Those Obtained in Simulations}}\label{subsec: sim-based auto-tuning}

\begin{figure}[b]
    \centering
    \includegraphics[width = \columnwidth]{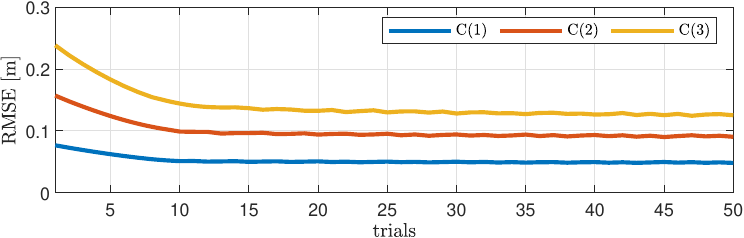}
    \caption{\revision{Simulation-based auto-tuning performance for a quadrotor (with identical physical parameters to the real quadrotor in experiments) on circular trajectories with speeds ranging from 1 to 3 m/s.)}}
    \label{fig:sim tuning RMSE reduction}
\end{figure}

\revision{
In this subsection, we compare the parameters auto-tuned in experiments (detailed in Section~\ref{subsec: tuning on three trajectories}) with those obtained in simulations. For the latter, we apply Algorithm~\ref{algo: full algorithm using sensitivity propatation} to a quadrotor system in simulation, where the vehicle's physical parameters (mass and inertia) are identical to the quadrotor used in the experiments. Furthermore, to stay consistent with the setup in experiments, we use the same initial controller parameters, loss function, horizon of evaluation, gradient clipping, and sampling time as in Section~\ref{subsec: tuning on three trajectories}. Since equipment wear and tear is not an issue in simulation, we raise the auto-tuning budget to 50 trials and reduce the learning rate $\alpha$ to 0.01. We auto-tune parameters for the circular trajectories at speeds of 1, 2, and 3 m/s. In Fig.~\ref{fig:sim tuning RMSE reduction}, we show the tracking errors through the trials. The tracking error shows a monotone reduction in the beginning and ends with small oscillations when it terminates after 50 trials, 
% \textcolor{red}{One can observe a monotone reduction in the tracking error in the beginning, and the loss curve ends with small oscillations when the 50 trials terminate,} 
which indicates that the best performance is achieved with the selected learning rate. We deploy the parameters auto-tuned in simulations on the real quadrotor used in the previous experiments in Section~\ref{subsec: tuning on three trajectories}. Furthermore, we test the parameters obtained in the 10th, 30th, and 50th trials in the simulation to examine the performance at different stages of the auto-tuning. The results are shown in Table~\ref{tb: exp comparison with sim}. In general, one can observe the reduced tracking error from the parameters that have been obtained through more trials in simulation, for example, with trajectories $C(1)$ and $C(2)$. 
However, the crashes seen on trajectories $C(2)$ and $C(3)$ with relatively more simulation trials indicate the common issue of sim-to-real gap: as the parameters evolve on simulation data, they inevitably (over)fit the simulation rather than the real system. 
The new results indicate the benefit of experiment-based auto-tuning, which is conditioned on the model-based gradient on data from a real system, and thus provides the best knowledge of parameter change for performance improvement.
Nevertheless, the parameters extracted from the 10th trial in simulations result in tracking errors close to the errors with the hand-tuned parameters, which may be used as initial parameters or warm start for further performance improvement in experiment-based auto-tuning.
}

\setlength{\tabcolsep}{5pt} % Default value: 6pt
\renewcommand{\arraystretch}{1} % Default value: 1
  \captionsetup{%size=footnotesize,
	%justification=centering, %% not needed
	skip=5pt, position = bottom}
\begin{table}[t]
	\centering
	\small
	\revision{
	% \vspace{-0.2cm}
	\captionsetup{font=small}
	\caption{\revision{Tracking RMSE (unit: [cm]) of the real quadrotor using simulation-based auto-tuning parameters (Sim$_{X}$), experiment-based auto-tuning parameters (ExpT), and hand-tuned parameters (HT). The subscript $X$ in Sim$_{X}$ indicates the number of trials in the simulation auto-tuning. For ExpT, parameters P1, P2, and P3 from Table~\ref{tb: parameters summary for quadrotor experiment} are applied to trajectories $C(1)$, $C(2)$, and $C(3)$, respectively. The hand-tuned parameters can be found in Table~\ref{tb: parameters summary for quadrotor experiment}.}}
	\begin{tabular}{l|ccccc}
		\toprule[1pt]
        Traj. & Sim$_{10}$ & Sim$_{30}$ & Sim$_{50}$ & ExpT & HT \\
        \midrule
        $C(1)$ & 7.3 & 6.6 & \textbf{4.3}  & 5.7 & 9.0\\
        $C(2)$ & 18.2 & 16.2 & crash  & \textbf{10.7} & 18.0\\
        $C(3)$ & 32.3 & crash & crash  & \textbf{15.4} & 28.6\\
       	\bottomrule[1pt]
	\end{tabular}
 \label{tb: exp comparison with sim}
	% \vspace{-0.5cm}
 }
\end{table}
\normalsize

% logs 573-585, Jan. 15, 2023

\section{Conclusion} \label{sec: conclusion}
In this paper, we propose DiffTune: an auto-tuning method using auto-differentiation, with the advantage of stability, compatibility with data from physical systems, and efficiency. Given a performance metric, DiffTune gradually improves the performance using gradient descent, where the gradient is computed using sensitivity propagation that is compatible with physical systems' data. We also show how to use $\mathcal{L}_1$AC to mitigate the discrepancy between the nominal model and the associated physical system when the latter suffers from uncertainties. Simulation results (on a Dubin's car and a quadrotor) and experimental results (on a quadrotor) both show that DiffTune can efficiently improve the system's performance. When uncertainties are present in a system,  $\mathcal{L}_1$ adaptive control facilitates tuning by compensating for the uncertainties.
Generalization of the tuned parameters to unseen trajectories during tuning is also illustrated in both simulation and experiments. 

One limitation of the proposed approach is that it only applies to systems with differentiable dynamics and controllers. The requirement on differentiability is not met in contact-rich applications~\cite{parmar2021fundamental} (e.g., legged robots~\cite{wieber2016modeling} and dexterous manipulation~\cite{jin2022task}) and systems with actuation limits~\cite{kumar2021diffloop} (e.g., saturations in magnitude or changing rate). Although subgradients~\cite{boyd2003subgradient} generally exist at the points of discontinuity, the impact of surrogate gradient on the tuning efficiency is unknown, which will be investigated in the future. \revision{Another limitation of this work is the convergence to a local minimum due to the usage of gradient descent. Future work will investigate conditions for convergence to a global minimum or methods that can help escape from local minimums.}

\section*{Acknowledgement}
The authors would like to thank Pan Zhao, Aditya Gahlawat, and Zhuohuan Wu for their valuable feedback during insightful discussions and Junjie Gao for his help with demo filming.

\section*{Appendix}
\subsection{Details of the Dubin's car simulation}\label{appdix: dubin car simulation}

\begin{figure}
    \centering
    \includegraphics[width = \columnwidth]{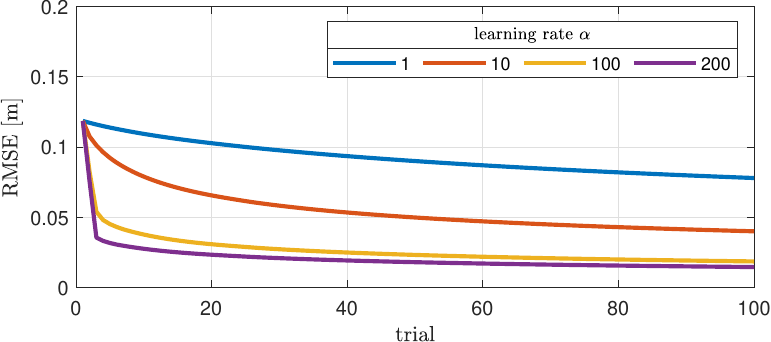}
    \caption{Performance of DiffTune in Dubin's car example with different learning rates.}
    \label{fig:Dubin's ar different learning rate}
\end{figure}

\noindent\textbf{Auto-tuning setup:} 
For Autotune~\cite{loquercio2022autotune}, we use a Gaussian distribution with variances set to 1.0 to sample the four parameters iteratively. The scoring function is the exponential of the tracking RMSE. Since sampling is used in AutoTune, we conduct 10 runs and show the mean, max, and min RMSEs in Fig.~\ref{fig: Dubin car comparison to Autotune and SafeOpt}. For SafeOpt~\cite{berkenkamp2016safe}, we choose the radial basis function kernel for GP. \revision{Both the length-scales and the kernel variance are set to 1.0. The noise variance for the model is set to 0.01. The objective function to maximize is negative tracking RMSE, and the safety threshold is set to $-1$ m. For GIBO~\cite{GIBO}, stochastic gradient descent is used as an optimizer, and the learning rate is set to 1. Like SafeOpt, the objective function to maximize is the negative tracking RMSE. The GP model uses a squared exponential kernel, and we assume the initial loss as a prior mean of the model. At each trial, four samples are used for a gradient estimate, and the last 20 sampled points are used for the approximation of the posterior. The normalized gradient is used for the parameter update.} For DiffTune, we use a learning rate {$\alpha=100$}. (The tracking RMSE's reduction with different learning rates in Dubin's car example is shown in Fig.~\ref{fig:Dubin's ar different learning rate}, which is consistent with how learning rate influences the loss reduction in general in a gradient-descent algorithm.) The feasible set of each parameter is set to $[1, 20]$ and applies to all three methods compared.

For the batch auto-tuning for generalization, we see the loss function as the squared norm of the position tracking error summed over a horizon of 10 s and set the learning rate $\alpha$ to 0.1. The termination condition is the relative reduction in the total loss between two consecutive steps being smaller than 1e-4 of the current loss value. 

\begin{figure}
    \centering
    \includegraphics[width = \columnwidth]{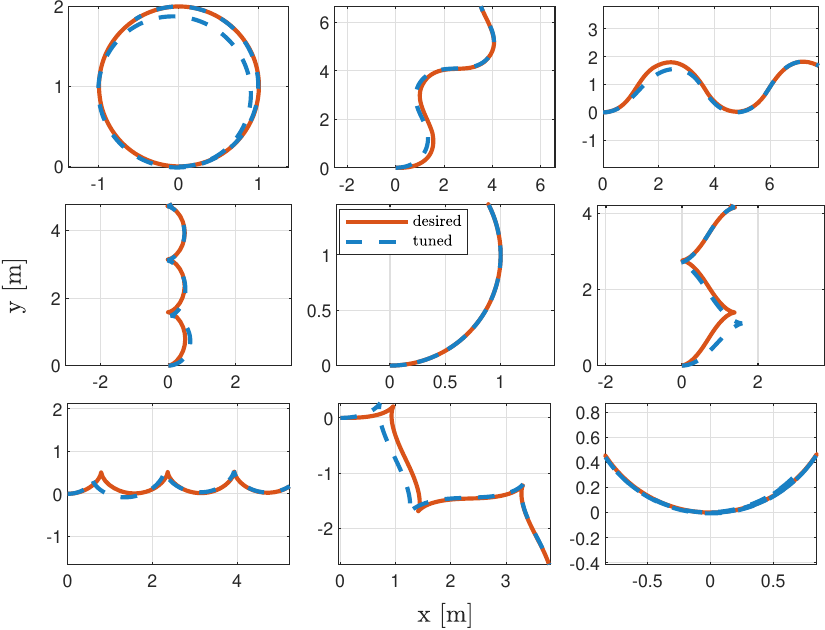}
    \caption{Trajectories for batch tuning in Dubin's car example.}
    % generated by dubinsCar_PDcontrol_fwdPropagation_newTraj_train9.m
    \label{fig:Dubins car training trajectories}
\end{figure}

The trajectories used for batch tuning for Dubin's car are shown in Fig.~\ref{fig:Dubins car training trajectories}, where the tuned parameters can achieve acceptable tracking on these trajectories.

\subsection{Details of the quadrotor simulation}\label{appdix: quadrotor simulation}

\setlength{\tabcolsep}{5pt} % Default value: 6pt
\renewcommand{\arraystretch}{1} % Default value: 1
  \captionsetup{%size=footnotesize,
	%justification=centering, %% not needed
	skip=5pt, position = bottom}
\begin{table*}[!b]
	\centering
	\small
	
	\vspace{-0.2cm}
	\captionsetup{font=small}
	\caption{Comparison of the initial parameters, DiffTune final parameters (P1, P2, and P3), DiffTune-$\mathcal{L}_1$AC final parameters (P1-$\mathcal{L}_1$, P2-$\mathcal{L}_1$, and P3-$\mathcal{L}_1$), and hand-tuned parameters}
	\begin{tabular}{l|rrr|rrr|rrr|rrr}
		\toprule[1pt]
		Param. & \multicolumn{3}{c|}{$\boldsymbol{k}_{\boldsymbol{p}}$} & \multicolumn{3}{c|}{$\boldsymbol{k}_{\boldsymbol{v}}$} & \multicolumn{3}{c|}{$\boldsymbol{k}_{R}$} & \multicolumn{3}{c}{$\boldsymbol{k}_{\boldsymbol{\Omega}}$}
		\\
		\midrule
	axis & $x$ & $y$ & $z$ & $x$ & $y$ & $z$ & $x$ & $y$ & $z$ & $x$ & $y$ & $z$ \\
 \midrule
 initial & 10 & 10 & 10 & 2 & 1 & 1.5 & 0.5 & 0.3 & 0.14 & 0. 11 & 0.1 & 0.01 \\
 \midrule
 P1 & 10.25 & 10.20 & 11.72 & 0.97 & 0.35 & 0.85 & 1.02 & 0.29 & 0.05 & 0.05 & 0.10 & 0.03 \\
P2 & 12.91 & 12.92 & 13.75 & 0.96 & 0.48 & 0.72 & 0.97 & 0.58 & 0.07 & 0.05 &  0.05 & 0.01
 \\
 P3 & 15.56 & 16.62 & 14.54 & 4.01 & 2.59 & 0.52 & 1.16 &  0.78 & 0.05
 & 0.07 & 0.03 &  0.03 \\
 \midrule
 P1-$\mathcal{L}_1$ & 10.52 & 10.44 & 10.04 & 0.71 &  0.34 & 1.40 & 1.15 & 0.67 & 0.13 & 0.06 & 0.04 & 0.03
 \\
 P2-$\mathcal{L}_1$ & 11.79 & 11.77 & 12.14 & 0.96 & 0.48 & 0.72 & 0.97 & 0.58 & 0.22 & 0.05 & 0.05 & 0.01
 \\
 P3-$\mathcal{L}_1$ & 16.29  & 16.11 & 16.26 & 1.27 & 0.95 & 0.52 & 1.29 & 0.77 & 0.05 & 0.07 & 0.04 & 0.03
 \\
 \midrule
hand-tuned & 14 & 15 & 15 & 1.50 & 0.90 & 1.10 &  0.55 & 0.35 &  0.15 &  0.04 & 0.03 & 0.01                  
\\
       	\bottomrule[1pt]
	\end{tabular}\label{tb: parameters summary for quadrotor experiment}
	% \vspace{-0.5cm}
\end{table*}
\normalsize

\noindent\textbf{Auto-tuning setup}:
\revision{For all methods, we use the same feasible set of parameters and the same horizon for performance evaluation.}
For AutoTune~\cite{loquercio2022autotune}, the variances of Gaussian distribution in the transition model are set to 2 for $\boldsymbol{k}_{\boldsymbol{p}}$, $ \boldsymbol{k}_{\boldsymbol{v}}$, and $\boldsymbol{k}_{{R}}$ and 1 for $\boldsymbol{k}_{\boldsymbol{\Omega}}$. The scoring function is the exponential of the RMSE tracking error.
{For the SafeOpt-PSO~\cite{duivenvoorden2017constrained}, the swarm size is set to 400 and we choose Matérn kernel
with parameter $\nu = 5/2$. The length-scales are set to 5 for $\boldsymbol{k}_{\boldsymbol{p}}$, $ \boldsymbol{k}_{\boldsymbol{v}}$, and $\boldsymbol{k}_{{R}}$ and 0.1 for $\boldsymbol{k}_{\boldsymbol{\Omega}}$. The kernel variance is set to 0.01, and the noise variance for the model is set to 0.05 for all trajectories. The objective function to maximize is negative tracking RMSE, and we set the safety threshold to $-1$ m. 
\revision{For GIBO~\cite{GIBO}, we have used the same setup for all trajectories. Stochastic gradient descent is used as an optimizer, and the learning rate is scheduled to 0.5 for the first 40 trials and then reduced to 0.2 until the maximum number of trials is reached. The objective function is set to the negative tracking RMSE (for maximization). The GP model uses a squared exponential kernel, and we assume the initial loss of the 3D Figure 8 trajectory as a prior mean of the model. We use a uniform prior between 0.1 and 5 as a hyper-prior of the length-scales. At each trial, 12 samples are used for a gradient estimate, and the last 12 sampled points are used to approximate the posterior. We set $\delta=0.5$ to limit local search bounds.
% , and all parameters are clamped above 0.5 to prevent them from turning negative. 
The normalized gradient is used for the parameter update.}

For DiffTune, the learning rate is set to 0.1. 
In addition, we tested three learning rates of $\alpha \in \{0.1,\ 0.01,\ 0.001\}$ for DiffTune. The results are shown in Fig.~\ref{fig:quadrotor_sim_loss_curve diffTune only}, where the rate of loss reduction is positively correlated to the magnitude of the learning rate. Furthermore, oscillation in the loss value is observed when $\alpha$ is relatively large (0.1), indicating the learning rate is too large when the parameters are close to the (local) minimum. These observations are consistent with how the learning rate influences the loss reduction in gradient descent.

In the ablation study of $\mathcal{L}_1$AC and DiffTune, we relax the upper bound on the feasible parameters so that the parameters can grow to higher values to handle the uncertainties.

\begin{figure}
    % \centering
    % \includegraphics[width = \columnwidth]{figures/loss_curve_lr.png}
    % \vspace*{-0.5cm}
    \begin{subfigure}{\columnwidth}
        \includegraphics[width = \columnwidth]{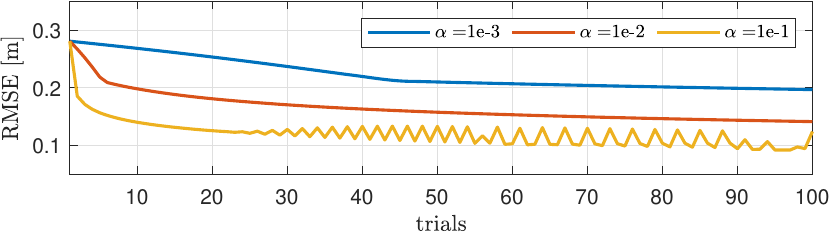}
        \caption{2D circle}
        \label{fig: 2D circle simulation RMSE}
    \end{subfigure}
    \begin{subfigure}{\columnwidth}
        \includegraphics[width = \columnwidth]{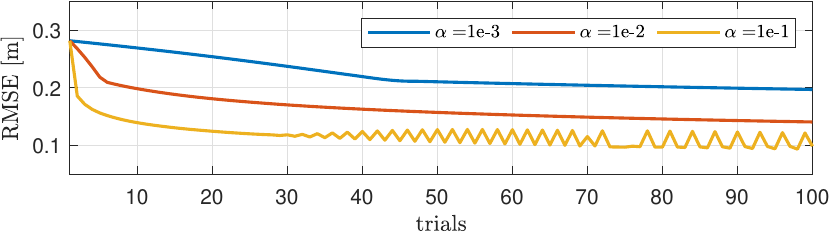}
        \caption{3D circle}
        \label{fig: 3D circle simulation RMSE}
    \end{subfigure}
    \begin{subfigure}{\columnwidth}
        \includegraphics[width = \columnwidth]{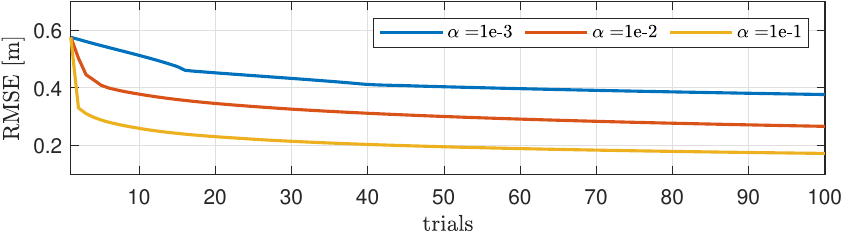}
        \caption{3D figure 8}
        \label{fig: 3D figure 8 simulation RMSE}
    \end{subfigure}
    \caption{Loss reduction by DiffTune under different learning rates $\alpha$ in the quadrotor simulation.}
  \label{fig:quadrotor_sim_loss_curve diffTune only}
\end{figure}

\subsection{Details of the quadrotor experiment}\label{appdix: quadrotor experiment}

\begin{figure}[t]
\begin{subfigure}{0.32\columnwidth}
    \centering
    \includegraphics[width = \textwidth]{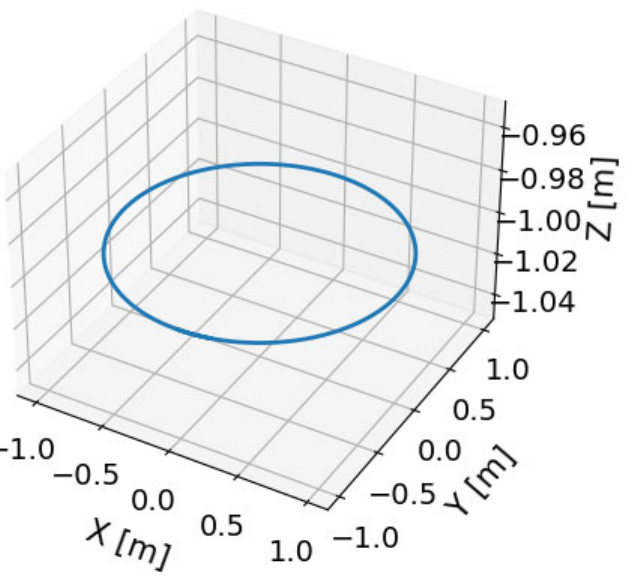}
    \caption{Circle}
    \label{fig: circular trajectory}
\end{subfigure}
\begin{subfigure}{0.32\columnwidth}
    \centering
    \includegraphics[width = \textwidth]{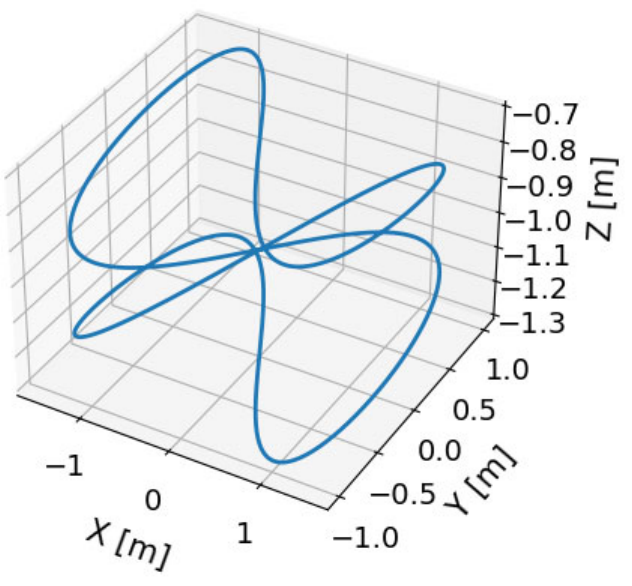}
    \caption{3D figure 8}
    \label{fig: 3D figure 8 trajectory}
\end{subfigure}
\begin{subfigure}{0.32\columnwidth}
    \centering
    \includegraphics[width = \textwidth]{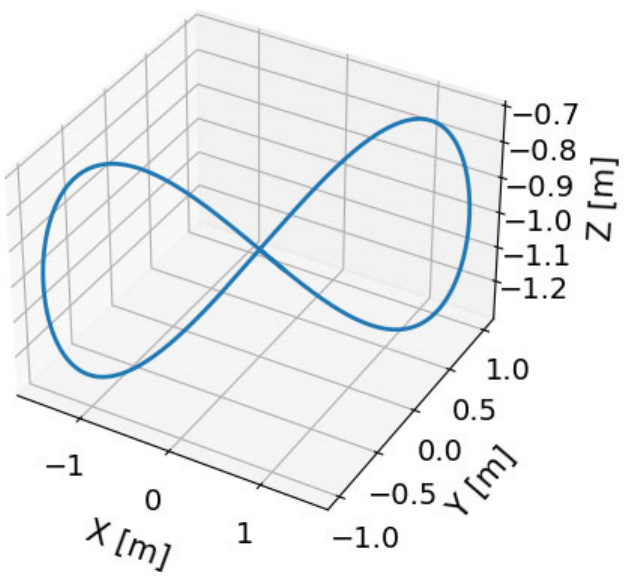}
    \caption{Vertical figure 8}
    \label{fig: vertical figure 8 trajectory}
\end{subfigure}
\caption{Trajectories used in the generalization test.}
\label{fig: trajectory shape for generalizability}
\end{figure}

We use a custom-built quadrotor to conduct the experiments. The quadrotor weighs 0.63~kg with a 0.22~m diagonal motor-to-motor distance. The quadrotor is controlled by a Pixhawk 4 mini flight controller running the ArduPilot firmware. We modify the firmware to enable the geometric controller and the $\mathcal{L}_1$ adaptive control, which both run at 400~Hz on the Pixhawk. Position feedback is provided by 9 Vicon V16 cameras. We use ArduPilot's EKF to fuse the Vicon measurements with IMU readings onboard.

The tuned parameters, without and with $\mathcal{L}_1$AC in the loop, are shown in Table~\ref{tb: parameters summary for quadrotor experiment}. The trajectories used for testing generalization in Table~\ref{tb: trajectories used for generalizability} are shown in Fig.~\ref{fig: trajectory shape for generalizability}.

\bibliographystyle{IEEEtran}
\bibliography{ref}
\end{document}